\documentclass{article}

\usepackage{arxiv}

\usepackage[utf8]{inputenc} 
\usepackage[T1]{fontenc}    
\usepackage{hyperref}       
\usepackage{url}            
\usepackage{booktabs}       
\usepackage{amsfonts}       
\usepackage{nicefrac}       
\usepackage{microtype}      
\usepackage{graphicx}
\usepackage[numbers,sort&compress]{natbib}
\usepackage{doi}
\usepackage{multirow}
\usepackage{pslatex}
\usepackage{amsmath}
\usepackage{amssymb}
\usepackage{mathrsfs}
\usepackage{xcolor}
\usepackage{textcomp}
\usepackage{manyfoot}
\usepackage{algorithm}
\usepackage{algorithmicx}
\usepackage[noend]{algpseudocode}
\usepackage{listings}
\usepackage{multicol}
\usepackage{subcaption}
\newcommand{\qdist}[1]{\textlangle#1\textrangle}

\begin{document}

\title{Introducing User Feedback-based Counterfactual Explanations (UFCE)\thanks{This is a preprint of a paper submitted to International Journal of Computational Intelligence Systems.}}


\author{ \href{https://orcid.org/0000-0002-1946-285X}{\includegraphics[scale=0.06]{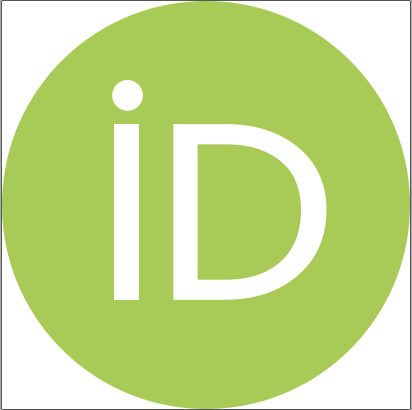}\hspace{1mm}Muhammad Suffian}
\\
	DISPeA\\
	University of Urbino\\
	Urbino, Italy \\
	\texttt{m.suffian@campus.uniurb.it} \\
	\And
        \href{https://orcid.org/0000-0003-3673-421X}{\includegraphics[scale=0.06]{orcid.png}\hspace{1mm}Jose M.~Alonso-Moral} \\
	CiTIUS\\
	Universidade de Santiago de Compostela\\
	Santiago de Compostela, Spain \\
	\texttt{josemaria.alonso.moral@usc.es} \\
        \And
	\href{https://orcid.org/0000-0001-6666-3315}           
        {\includegraphics[scale=0.06]{orcid.png}\hspace{1mm}Alessandro Bogliolo} \\
	DISPeA\\
	University of Urbino\\
	Urbino, Italy \\
	\texttt{alessandro.bogliolo@uniurb.it} \\
}

\date{}

\renewcommand{\shorttitle}{UFCE}

\hypersetup{
pdftitle={UFCE},
pdfsubject={CS},
pdfauthor={Muhammad Suffian},
pdfkeywords={XAI},
}

\maketitle

\begin{abstract}
Machine learning models are widely used in real-world applications. However, their complexity makes it often challenging to interpret the rationale behind their decisions. Counterfactual explanations (CEs) have emerged as a viable solution for generating comprehensible explanations in eXplainable Artificial Intelligence (XAI). CE provides actionable information to users on how to achieve the desired outcome with minimal modifications to the input. 
However, current CE algorithms usually operate within the entire feature space when optimizing changes to turn over an undesired outcome, overlooking the identification of key contributors to the outcome and disregarding the practicality of the suggested changes.
In this study, we introduce a novel methodology, that is named as user feedback-based counterfactual explanation (UFCE), which addresses these limitations and aims to bolster confidence in the provided explanations. UFCE allows for the inclusion of user constraints to determine the smallest modifications in the subset of actionable features while considering feature dependence, and evaluates the practicality of suggested changes using benchmark evaluation metrics. We conducted three experiments with five datasets, demonstrating that UFCE outperforms two well-known CE methods in terms of \textit{proximity}, \textit{sparsity}, and \textit{feasibility}. Reported results indicate that user constraints influence the generation of feasible CEs.
\end{abstract}


\section{Introduction}
\label{sec:intro}
Nowadays, black-box machine learning (ML) models are extensively employed in different applications that frequently have an impact on human lives (e.g., lending, hiring, insurance, or access to welfare services) \cite{ijcisfinance,ajunwa2016hiring, dexe2020empirical}. In this context, understanding and trustworthiness of ML models are crucial. However, many ML models' internal workings appear to be opaque. Several questions raised to scrutinise and debug their behaviour usually remain unanswered: e.g., Why did we receive this result? What changes could provide an alternative outcome?  

When ML systems engage humans in the loop, they are expected to meet at least one of two requirements: (1) explain model prediction and (2) provide helpful suggestions for assisting humans to achieve their desired outcome \cite{kulesza2013too, o2020explainable, suffianHcxaiSurvey}.
For example, consider a Bank Loan application problem in which a user asks for a loan from an online banking service and the decision is made automatic by an intelligent agent. Such a decision (classification) can be challenged in the case of unfavourable outcomes from the loan applicant. Therefore, the loan applicant should be provided with an explanation of the factors involved in the classification and suggestions about how to change an unfavourable outcome. These two requirements are fulfilled with factual and counterfactual explanations in the field of Explainable Artificial Intelligence (Explainable AI or XAI\footnote{XAI stands for eXplainable Artificial Intelligence. This acronym became popular when the USA Defence Advanced Research Projects Agency (DARPA) introduced the challenges of designing self-explanatory AI systems \cite{gunning2021darpa}.} for short) \cite{ISJ-shao2023effect, stepin2021survey, ISJ-ding2022explainability, adadi2018peeking, stepin2021factual}. Factuals refer to what is observed in the actual scenario (e.g., ranking the most important input factors \cite{ribeiro2016should,ISJ-zhou2020many, lundberg2017unified}). In contrast, counterfactuals refer to simulated imaginary scenarios (e.g., increased income could provide alternative outcomes \cite{wachter2017counterfactual}) in the application domain \cite{judea-bookofwhy}. These hypothetical scenarios could provide information similar to the original input as \textit{``specifying necessary minimal changes in the input so that a favourable outcome is obtained''}, what is also called a \textit{Counterfactual Explanation} (CE) \cite{stepin2021survey}. It is worth noting that CE has been deemed acceptable for the General Data Protection Regulation (GDPR) in the European Union \cite{voigt2017eu}.

Wachter et al. \cite{wachter2017counterfactual} proposed one of the earliest methods for generating CEs, which involves adjusting input features to achieve a desired outcome such as loan approval. Although, various approaches have been proposed subsequently \cite{mothilal2020explaining, AR-survey, JMA-multimodal, ISJ-fernandez2022explanation}, they do not provide human-centred explanations (e.g., explanations containing actionable information grounded in user requirements). CEs generated by current approaches may recommend impractical actions (e.g., extreme input modifications) due to the lack of consideration for user feedback.  

User feedback can contribute to address the above mentioned problems, and thus improve the generation of meaningful and actionable explanations. Accordingly, we propose the User Feedback-based Counterfactual Explanations (UFCE) algorithm, which allows the user to specify the scale of input modifications to generate CE recommending actionable information, consequently providing viable outcomes. In a previous work \cite{fce, suffianieeecim}, the authors already dealt with the notion of feedback while their scope of user involvement was limited to defining the neighbourhood to minimise the \textit{proximity} constraint. Here, we extend previous work by expanding its scope to find out the \textit{mutual information} (see Section \ref{MI}) of key contributors in the feature space and considering user constraints to define a feasible \textit{subspace}\footnote{The meaning of subspace is equivalent to subset in our context, and it should not be confused with subspace in mathematics.} (see Section \ref{nn}) to search CEs. The proposed algorithm guides the search in this subspace to find the minimal changes (\textit{perturbations}) to the input that can alter the classification result as required. UFCE introduces three methods to perform minimal changes in the input features (see Section \ref{perturbations}), i.e., suggesting \textit{single feature}, \textit{double feature} and \textit{triple feature} changes in the input at a time). The adherence of minimal changes in the input features to the \textit{subspace} confirms the suggested actions as feasible. The mutual information of features guides in deciding which features to perturb (see Section \ref{perturbations}).
UFCE is a deterministic, model-agnostic, and data-agnostic approach for tabular datasets. In this paper, the focus is on a binary classification problem while the extension of UFCE for  multi-class classification problems will be addressed in future work.

In addition, UFCE is subjected to a rigorous evaluation focussing on \textit{sparsity}, \textit{proximity}, \textit{actionability}, \textit{plausibility} and \textit{feasibility} of explanations.
Compared to existing solutions, UFCE not only demonstrates enhanced performance in terms of subjected evaluation metrics but also makes a substantial contribution in the current state of the art.

\begin{figure}[h!]
    \centering
    \includegraphics[width=.7\textwidth]{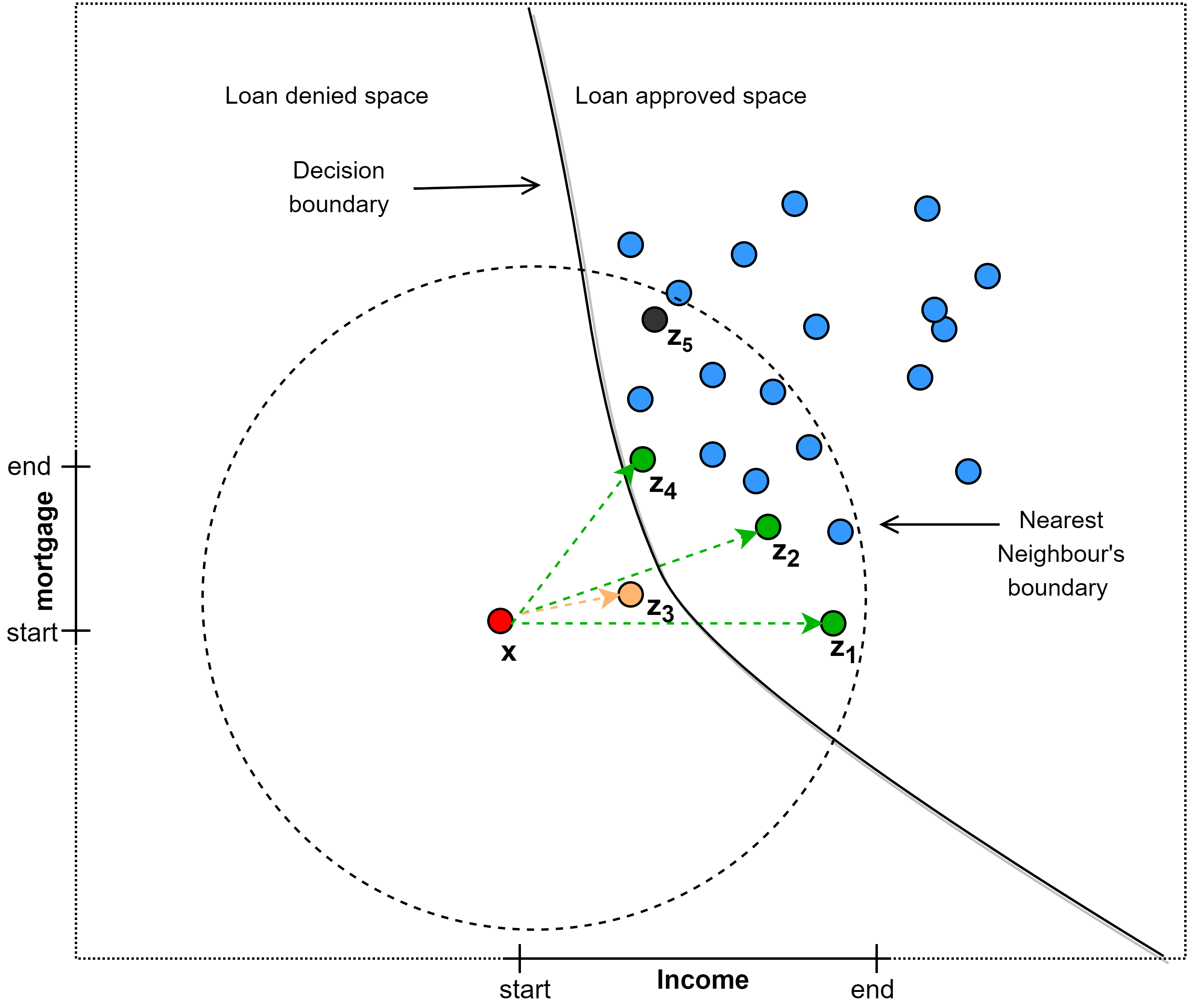}
    \caption{Example of decision surface with counterfactual instance space in the neighbourhood of test instance $x$. The yellow, black, and green dots ($z_1$, $z_2$, $z_3$, $z_4$, $z_5$) are the counterfactual instances: where $z_3$ is invalid; $z_1$, $z_2$, and $z_4$ are valid and actionable; and $z_5$ is valid but not actionable due to not adhering to user defined feature range for Mortgage (assume \textit{Bank loan} data).}\label{decision-surface-cfs}
\end{figure}

In Fig.~\ref{decision-surface-cfs}, an illustrative example of counterfactual instance space is shown. The decision boundary divides this space into `loan denied space' and `loan approved space'. The blue dots (points) represent the instances with the outcome of loan approval in the actual space of Loan data, and the red dot ($x$) represents an instance with the outcome of a loan denied (test instance). The green, black, and yellow counterfactual instances ($z_1$, $z_2$, $z_3$, $z_4$, and $z_5$) are produced due to the smallest feature modifications to $x$ in its \textit{nearest neighbours}. The nearest neighbourhood is shown with a dotted curved line on the decision boundary in the loan-approved space 
(see Section \ref{nn}). The instance $z_3$ represents inadequate changes in $x$ that could not result in its outcome to loan approved, whereas $z_1$, $z_2$, $z_4$, and $z_5$ represent sufficient changes in $x$ resulting in their outcomes to loan approved. In addition, $z_1$, $z_2$, and $z_4$ adopt the changes in mutually informed features adhering to the subspace which guarantees them as feasible counterfactuals, and convincing the model to alter their outcomes (i.e., desired outcomes), whereas $z_5$ does not adhere to the user specified feasible ranges of features making it in-actionable (unfeasible), while $z_3$ could not convince the prediction model to alter its outcome. For the sake of explaining in-actionable counterfactual $z_5$, we assumed 2D plot labelled with start and end range of Income and Mortgage features specifying the actionable subspace by the user and $z_5$ violates the range of Mortgage making it not actionable and unfeasible. 

In summary, the main contributions in this paper are as follows:
\begin{enumerate}
    \item We propose the UFCE algorithm that can generate actionable explanations complying with user preferences in mixed-feature tabular settings.
    \item We provide experimental evidence by simulating different kinds of user feedback that an end user can presumedly provide. We observed that user feedback is influencing in obtaining feasible CEs, and UFCE turned out to be the most promising algorithm as compared to two well-known CE generation algorithms: Diverse Counterfactual Explanations (DiCE) and Actionable Recourse (AR).
    \item We analyse the proposed UFCE algorithm and evaluate its performance on five datasets in terms of widely used evaluation metrics, including sparsity, proximity, actionability, plausibility, and feasibility. We present the results obtained from a simulation-based experimental setup for UFCE, DiCE, and AR. We observe how UFCE outperforms DiCE and AR in terms of\textit{ proximity}, \textit{sparsity}, and \textit{feasibility}.
    \item We implemented our algorithm as open-source software in Python, and it is made publicly available to support further investigations.
\end{enumerate}

The rest of the paper is organised as follows: Section \ref{sec:rw} revisits the related work, including methods and frameworks. Section \ref{preliminaries} introduces the preliminaries, problem statement and evaluation metrics. Section \ref{ufce} introduces the new UFCE algorithm. Section \ref{experiments} details the experimental setting and discusses the reported results in the empirical study. Finally, Section \ref{conclusion} draws some conclusions and points out future work.

\section{Related Work}\label{sec:rw}
\begin{table*}[hbt!]
	\centering
	\caption{Categorisation of CE algorithms: Optimisations (OPT), Heuristic Search (HS), Decision-Tree (DT), Fuzzy-Rules (FR), Instance-Based (IB), and Instance and Neighbourhood based (IB-NB).}
	\label{tab:RW-taxonomy}
	\tabcolsep3pt 
	\begin{tabular}{p{0.20\textwidth}p{0.12\textwidth}p{0.1\textwidth}p{0.1\textwidth}p{0.1\textwidth}p{0.14\textwidth}p{0.1\textwidth}}
		\toprule
		Algorithm & Strategy & \begin{tabular}[c]{@{}c@{}}Model-\\agnostic\end{tabular} & \begin{tabular}[c]{@{}c@{}}Data\\ Type\end{tabular} & \begin{tabular}[c]{@{}c@{}}Mixed-\\Features\end{tabular}  & \begin{tabular}[c]{@{}c@{}}User\\Constraints\end{tabular} & Code \\\midrule
		Wachter et al.  & OPT & \checkmark   & TAB & \checkmark & - & -  \\
		DiCE            & OPT & \checkmark     & TAB & \checkmark & \checkmark &  \href{https://github.com/interpretml/DiCE}{link}\\
		FCE             & OPT & \checkmark     & TAB & \checkmark & \checkmark & \href{https://github.com/msnizami/FCE}{link}\\
		AR              & OPT & LR              & TAB & \checkmark & \checkmark & \href{https://github.com/ustunb/actionable-recourse}{link} \\
		CLEAR           & HS  & \checkmark      & TAB & \checkmark & - & \href{https://github.com/ClearExplanationsAI/CLEAR}{link}  \\
		GSG             & HS  & \checkmark      & ALL & \checkmark & -  & \href{https://github.com/thibaultlaugel/growingspheres}{link} \\
		LORE            & DT  & \checkmark      & ALL & \checkmark & - &  \href{https://github.com/riccotti/LORE}{link} \\
		Stepin et al.   & FR  & FRBCS           & TAB & -          & - &  \href{https://gitlab.citius.usc.es/ilia.stepin/fcfexpgen/-/tree/master}{link} \\
		FACE            & IB  & \checkmark      & ALL & \checkmark & - & - \\
		NNCE            & IB  & \checkmark      & TAB & \checkmark & - & \href{https://pair-code.github.io/what-if-tool/explore/}{link} \\
		UFCE (this paper) & IB-NB & \checkmark   & TAB & \checkmark & \checkmark & \href{https://github.com/msnizami/UFCE}{link} \\ \bottomrule
	\end{tabular}
\end{table*} 

XAI has witnessed substantial growth over the last decade. Our research focuses on CEs as a means to explain AI. Consequently, we restrict our investigation to CE research in the subsequent paragraphs. For a comprehensive understanding of XAI and its existing algorithms, we recommend referring to Ali \emph{et al.} \cite{jose2023wwwxait} and Holzinger \emph{et al.} \cite{b3}.
This section briefly reviews the CE generation algorithms related to our work. The selected papers discussed below are purposefully chosen as they closely align with our approach.
Watcher \emph{et al.} \cite{wachter2017counterfactual} introduce the preliminary idea of CEs and evaluate their compliance with regulations. They frame the process of generating counterfactuals as an optimisation problem that seeks minimal distance between two data points by minimising an objective function using gradient descent. Their principal objective is to determine the proximity of data points, which is evaluated using a pertinent distance metric applicable to the dataset, such as L1/L2 or customised distance functions.
The DiCE algorithm \cite{mothilal2020explaining} emphasises feasibility and diversity of explanations, which can be optimised by applying the gradient descent algorithm to a loss function. DiCE assumes feature independence during perturbations for counterfactual generation. Yet, real-world features often correlate and hold mutual information, challenging the universal applicability of feature independence assumption.
Feedback-based Counterfactual Explanation (FCE) \cite{fce} is built upon the strengths of the state-of-the-art explanatory method given by Wachter \emph{et al.} \cite{wachter2017counterfactual}. FCE focused on defining a neighbourhood space around the instance of interest with user feedback and finds minimal distant CE in its proximity that provides favourable outcomes.   
Ustun \emph{et al.} initially resolved the issue of actionability in CE generation by introducing the AR algorithm \cite{ar}, which can handle categorical features by discretising numerical features; however, discretisation could be a weakness since it encodes feature values into a different format and decoding new values is likely to yield a precision problem after perturbations. The mixed integer programming model optimises the recourse to comply with restrictions prohibiting immutable features from being modified.
 
Counterfactual Local Explanations via Regression (CLEAR) \cite{clear} is based on heuristic search strategies to discover CEs using local decisions that minimise a specific cost function at each iteration. Firstly, CLEAR explains single predictions through ``boundary counterfactuals'' (b-counterfactuals) that specify the minimal adjustments required for the observation to ``flip'' the output class in the case of binary classification. Secondly, explanations are created by building a regression model that aims to approximate the local input-output behaviour of the ML system. Growing Spheres (GS) \cite{gsg} is based on a generative algorithm that expands a sphere of artificial instances around the instance of interest to identify the closest CE. Until the decision boundary of the classification model is crossed and the closest counterfactual to the instance of interest is retrieved. This algorithm creates candidate counterfactuals at random in all directions of the feature space without considering their feasibility, actionability and (un)realistic nature in practice.

Decision tree-based explainers uncover CE using the tree's structure to simulate the opaque ML model behaviour. These techniques first estimate the behaviour of a black-box model with a tree and then employ the tree structure to extract CE \cite{lore,stepin2022empirical}.
Local Rule-based Explainer (LORE) \cite{lore} is a decision-tree approach that uses factual and counterfactual rules to explain why a choice was made in a given situation. It starts by sampling the local data for the explanation using a genetic algorithm. Then, LORE uses the sampled neighbourhood records of a particular instance to train a decision tree, which supports the generation of an explanation in the form of decision rules and counterfactuals.
Another algorithm for factual and CEs \cite{stepin2022empirical} assess the length and complexity of rules in a fuzzy rule-based classification system (FRBCS) to estimate the conciseness and relevance of the explanations produced from these rules.

Feasible and Actionable CE (FACE) \cite{face2020} is an instance-based CE generation algorithm that extracts CEs from similar examples in the reference dataset. It accounts for the actionable explanations based on multiple data paths and follows some feasible paths achievable by the shortest distance metric defined on density-weighted metrics. Thus, FACE constructs a graph on the selected data points and applies the shortest distance path algorithm (Dijkstra's Algorithm) to find the feasible data points for generating CEs.
The Nearest-Neighbour CE (NNCE) \cite{nnce} is also an instance-based explainer that chooses the examples in the dataset which are the most similar to the instance of interest but associated with an output class different from the actual. The computational expense of calculating distances between input instances and every occurrence in a dataset with a different result is a shortcoming of FACE and NNCE methods. 

We categorised and summarised the above-mentioned CE algorithms in Table \ref{tab:RW-taxonomy}. The strategies are Optimisations (OPT), Heuristic Search (HS), Decision-Tree (DT), Instance-Based (IB), and a hybrid of Instance and Neighbourhood based (IB-NB). If it is a model-agnostic approach, we use a check mark (\checkmark); in other cases, the specific classifier is tagged. If it can process all types of data (ALL); otherwise, a specific data type is mentioned, like Tabular (TAB), Image (IMG), and Text (TXT). A \checkmark is used for handling mixed features (categorical and numerical) and user constraints. A link is provided on the availability of the source code.

Even if many alternative algorithms exist for CE generation, only DiCE, AR and FCE handle user constraints as an input, and they are therefore the only CE generation algorithms somehow comparable to the new algorithm that is proposed in this work. Accordingly, in the rest of this manuscript, we go deeper with designing and evaluating a novel CE generation algorithm enriched with user feedback.

\section{Preliminaries}\label{preliminaries}
In this section, we introduce the notation (i.e., expressions and terms) used throughout the paper (Sec.~\ref{notations}), the problem statement (Sec.~\ref{probstat}), and evaluation metrics for counterfactual explanations (Sec.~\ref{cf-properties}).

\subsection{Expressions and Terms}\label{notations}
In the field of XAI, different important notions and concepts are explained with multiple expressions and terms. The definitions of these expressions and terms cannot be universally applied to different counterfactual approaches in different contexts, and must be redefined first. The specific meanings of frequently used expressions and terms throughout the paper are presented in Table~\ref{tab:terms}.
\begin{table*}[t!]
\centering
\caption{The definitions of frequently used expressions and terms.}
\label{tab:terms}
\resizebox{\textwidth}{!}{%
\begin{tabular}{p{0.15\textwidth}p{0.8\textwidth}}
\toprule
Terms & Description \\ \midrule
Validity & A CE is valid if it changes the classification outcome with respect to the original one. \\ 
Feasibility & It is a combined property of any CE with a Boolean value (when CE is valid, plausible and actionable); we refer to it when CE follows all feasibility properties. \\ 
Sparsity & The percentage of feature changes between the test instance and the counterfactual. \\ 
User-preferences & It is similar in meaning to expressions like user-specified and user-constraints, representing the choice of features and restrictions to their values. \\ 
Actionability & The percentage of feature changes included within user-specified list of features that are actually applied to obtain CE. \\ 
Desired space & It is a subset of the training data which contains only positive outcome-based data instances.  \\ 
Subspace & It is a space created with the lower and the upper bounds of the features. \\ 
Plausible space & The original distribution of the data (an input space). \\ 
Plausibility & It refers to the adherence of CE to the plausible space. \\ \bottomrule
\end{tabular}%
}
\end{table*}

\subsection{Problem Statement}\label{probstat}
Let us assume we are given a point $x = (x_1,...,x_d$), where $d$ is the number of features. Each feature takes values either in (a subset of) $\mathbb{R}$, in which case we call it a numerical feature or in (a subset of) $\mathbb{N}$, in which case we call it a categorical feature (binary categories). For categorical features, we use natural numbers as a convenient way to identify their categories but disregard orders. For example, for the categorical feature `Online', $0$ might mean \emph{no}, and $1$ might mean \emph{yes}. Thus, $x \in \mathbb{R}^{d_1} \times \mathbb{N}^{d_2}$, where $d_1 + d_2 = d$.

A counterfactual instance\footnote{Many authors use $x'$ to denote it as a counterfactual point that confuses with prime of $x$ in mathematics. Thus, we are using another letter $z$ to denote the counterfactual point.} for a test instance $x$ is an instance $z \in \mathbb{R}^{d_1} \times \mathbb{N}^{d_2}$ such that given a ML (black-box) classification model $f: \mathbb{R}^{d_1} \times \mathbb{N}^{d_2} \rightarrow{y}$ and $y=\{0, 1\}$ is a decision or class (this study considers a binary classification task), and $f(x) \neq f(z)$ (we follow notations from \cite{virgolin2023robustness}).

Our approach addresses the assumption that features are either dependent or independent from each other to compute $z$, as it frequently happens in real-world practice. We handle these dependencies by exploiting the mutual information (MI) shared among the features and utilising it in the selection of features to perturb (see Section~\ref{MI}).

A CE is an intervention in $x$ that reveals how $x$ needs to be changed to obtain $z$. We will now examine the traditional setting in which we have multiple $z$; for the sake of clarity and without compromising generality, we seek the most suitable $z^*$ with:
\begin{align}\label{eq-1}
     \begin{array}{l}
     z^* = \operatorname*{argmin}_{z} \delta (z, x) \\
    \text{with} \quad \delta (z, x) = prox\_Jac(z, x) + \lambda \cdot prox\_Euc(z, x)  \\
     \text{subject to} \quad f(z)=t \quad \text{and $z$ is plausible.}
    \end{array}
\end{align}
where $f$ is a ML model, $t$ denotes the desired output and $\delta$ determines the distance. We wish $z$ to be close to $x$ under the distance function $\delta$ that handles categorical and numerical features as a linear combination of their categorical and numerical distances. The categorical distance $prox\_{Jac}$ is represented as following: 
\begin{equation} \label{eq:proxJac}
    prox\_Jac(z, x) = 1 - Jacc(z, x)
\end{equation}
where $prox\_Jac(z, x)$ measures the distance of categorical features using $Jacc(.,.)$ that represents a Jaccard index. The result of this mathematical notation is a value between $0$ and $1$, where a value of $0$ indicates that $x$ and $z$ are identical in terms of categorical features, and a value of $1$ indicates that all categorical features in $x$ are different from those in $z$.
The numerical distance $prox\_Euc(z,x)$ measures the Euclidean distance between $x$ and $z$ for numeric features. The parameter $\lambda$ balances the influence of the two distances (re-scaling factor). To make the addition of both distances possible on the same scale, we have to normalize the numerical distance in $[0, 1]$. 

\subsection{Evaluation metrics for Counterfactual Explanations}\label{cf-properties}
Evaluating the quality of CEs is an important task, as it helps to ensure that the explanations provided are accurate and informative. Here are defined the evaluation metrics that we will use to analyse the quality of CEs in the empirical study (Section \ref{experiments}).

Sparsity is defined as the average number of changed features in CE versus the test instance (also, the percentage of feature changes). This is desirable to take a small value because the user can often only reasonably focus on and intervene upon a limited number of features (even if this amounts to a higher total cost than intervention on all the features). The following notation 
($ \frac{1}{d}\sum_{i=1}^{d} 1\; s.t.\; z_{i} \neq x_{i}, 0$ otherwise) computes sparsity of $z$. Where $d$ is the number of features, $z_{i}$ and $x_{i}$ represent to the $i^{th}$ feature in $z$ and $x$, respectively.

Proximity is defined as the distance from $z$ to $x$. We measure two types of proximity, $prox\_Jac$ for categorical features (see Eq.~\ref{eq:proxJac}), and $prox\_Euc$ for numerical features (see Eq.~\ref{eq-euc}).

\begin{equation}\label{eq-euc}
    prox\_Euc(z,x)   = \sum _{i=1}^{d} \frac{ \sqrt{ \left( z_{i}-x_{i}\right)^2}}{MAD_i}    
\end{equation}

Using Euclidean distance, the Eq.~\ref{eq-euc} measures the distance $prox\_Euc(z, x)$ between $x$ and $z$ for numerical features weighted with median absolute deviation (MAD) of respective feature. The result of this equation is a value that represents the similarity between $x$ and $z$, with higher values indicating greater distance, i.e., smaller proximity. 

Plausibility evaluates whether CE is plausible or not (Boolean outcome). We used the Local Outlier Factor algorithm, which computes the local density deviation of a given data point with respect to its neighbours \cite{lof}, and it is merely one choice among several that may be used to assess the level of plausibility, like other outlier detection techniques (e.g., Isolation Forests) \cite{MD-lof-survey}.

Actionability counts for the average number of feature changes (also the percentage of feature changes) suggested from the user-specified list of features. It is a similar metric to sparsity; however, the feature changes are not counted for all features but rather for a subset of features (only for user-specified features).

Feasibility pays attention, at the same time, to validity, actionability, and plausibility. Validity is related to the accomplishment of the desired outcome ($t$) given $z$ such that $f(z)=t$. In addition, we use a certain threshold of feature changes in $z$, if fulfilled, to be admitted as an actionable counterfactual. The following equation measures the outcome of feasibility for $z$ (a Boolean outcome).
 \begin{align}
    \begin{array}{l}
     z_{feasible}=(z_{valid}=1) \And (z_{plausible}=1) \\ \And (z_{actionable} >= Actionable_{thresold})
     \end{array}
 \end{align}

Finally, we also record the required computational time, i.e., the average time taken (seconds) to generate one CE.

\section{The counterfactual generation pipeline}\label{ufce}
\begin{figure*}[hbt!]
    \includegraphics[width=\textwidth]{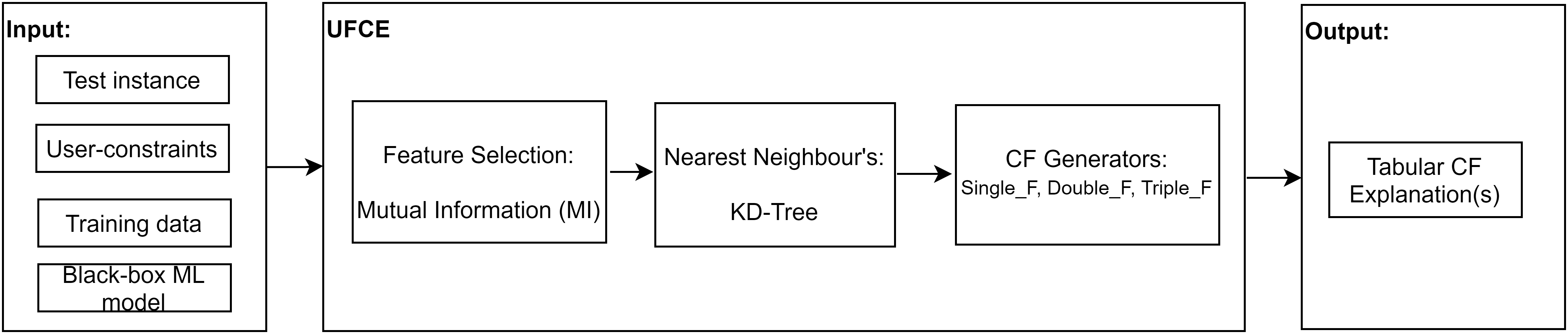}
    \caption{Counterfactual explanation generation pipeline of UFCE.}\label{ufce-pipeline}
\end{figure*}

Researchers have suggested several guidelines for the practical utility of CEs, including the minimal effort to align with user preferences \cite{dandl2020multi}. In this paper, we consider the interplay between the desiderata of user preferences (i.e., that CEs require only a subset of the features to be changed) and feasibility measures (i.e., that CEs remain feasible and cost-effective). 
The main building blocks and pipeline of UFCE are shown in Fig.~\ref{ufce-pipeline} and they will be described in detail in the rest of this section. The main components are input (a trained ML model on a dataset, a test instance to be explained along with user constraints), UFCE (counterfactual generation mechanism that includes feature selection, finding nearest neighbours, and calling the different perturbation methods for counterfactual generation), and output (tabular presentation of generated counterfactuals).

\subsection{User Preferences}\label{up}
A counterfactual instance might be close to being realised in the feature space. Still, due to limitations in the real world, it might only sometimes be feasible \cite{wachter2017counterfactual, sokol2020one, ehsan2021operationalizing}. Therefore, enabling users to impose constraints on feature manipulation is natural and intuitive. These constraints can be imposed in two ways: firstly, the user may specify features which can be modified; secondly, he/she can set the feasible range for each feature, within which the counterfactual instances must be located. For instance, a constraint could be ``income cannot exceed \$10,000''. Given the feature values that the user describes as a starting point, we seek minimal changes to those feature values that result in an instance for which the black-box model makes a different (often a specific favourable one) decision. 

A particular novelty of our approach is that it focuses on perturbing only the subset of features that are deemed as  relevant and actionable by the user. This novelty is based on selecting features that make a higher impact on the target outcome and adhere to user-defined constraints. Note that it could be presumed that the user-defined constraints are the thoughts embodied in feasible ranges of features; however, these requirements are necessary preconditions and do not ensure that the explanations are aligned with the users’ cognitive abilities \cite{cogxai_suffian, sundar2020rise}. It is worth noting that we evaluate CEs only on the defined quantitative evaluation metrics (see Section \ref{cf-properties}), because dealing with human grounded evaluation is out of the scope of this work.

\subsection{Mutual Information of features}\label{MI}
Different strategies and assumptions have been exploited in the literature for feature selection and feature dependencies to account for CEs \cite{ar, carla, discern}. We use Mutual Information (MI) of features.
MI stands out as a robust feature selection method capturing both linear and non-linear relationships within data. Unlike some linear approaches, it does not assume a specific data distribution, enhancing its versatility across diverse datasets. MI effectively identifies informative and non-redundant features, making it particularly valuable in scenarios with complex, non-linear relationships. Its robustness to outliers and applicability in high-dimensional spaces further contribute to its effectiveness as a feature selection tool.
The MI of two random variables, in probability theory and information theory \cite{shannon2001mathematical}, measures the extent of their mutual dependence. Unlike the correlation coefficient, which is restricted to linear dependence and real-valued random variables, MI can determine the degree of difference in the joint distribution of the two variables in a comprehensible way. It evaluates the ``amount of information'' about one variable that can be gained by observing the other variable. MI is closely related to the concept of entropy, which measures the expected ``amount of information'' held in a random variable and is a fundamental idea in information theory \cite{kolmogorov1956shannon}.
The MI of two random variables ($x_i$ and $x_j$) is represented in Eq.\ref{eq-mi} as following:
\begin{equation}\label{eq-mi}
I(x_i, x_j) =  H(x_i, x_j) - H(x_i|x_j) - H(x_j|x_i)
\end{equation}
where $H(x_i)$ and $H(x_j)$ are the marginal entropies, $H(x_i|x_j)$ and $H(x_j|x_i)$ are the conditional entropies, and $H(x_i, x_j)$ is the joint entropy of $x_i$ and $x_j$ ($x_i$ and $x_j$ are $i^{th}$ and $j^{th}$ variables). A zero value of $I(x_i, x_j)$ indicates that $x_i$ and $x_j$ are independent, while large values indicate great dependence. We compute MI by exploiting this functionality from the scikit-learn package in Python, which employs non-parametric techniques and uses entropy estimation derived from k-nearest neighbours distances, outlined in \cite{kraskov2011erratum}.

Regarding feature independence, the features are perturbed individually to a defined range. These perturbations are termed as \textit{single-feature perturbations}. 
We select tuples (double and triple features) based on their MI and sort them in descending order (from higher to lower MI) in case of regarding feature dependence. These sorted tuples also help to decide which one to use (the common with the user-specified features) for the perturbations. In the case of double features, the $I(.,.)$ function already provides the scores of MI for each pair of features from where we select the top pairs of features. To form a triplet, we use already computed top pairs of features making their triplet with each feature from the user-specified list of features (with no repetition of features in a triplet). The order of tuples is preserved to the actual distribution of data. The perturbation in these tuples (double and triple) features are termed as \textit{double feature perturbations} and \textit{triple feature perturbations}, respectively (see Section \ref{perturbations}). It is worth noting that for the case of triplet, we do not exploit the functionality of MI to find the causal relationship \cite{hilton1990conversational, holzinger2019causability, ISJ-baltsou2023explaining} that is out of the scope of this work.     

\subsection{Nearest Neighbourhood (NN)}\label{nn}
Under the problem statement that we considered in Sec.~\ref{probstat}, a hyper-rectangle, also known as a box, is formed by the potential perturbations that could affect a test instance $x$. This box encompasses all the possible paths to counterfactual $z$ that could be reached from $x$ due to these perturbations. In other words, these paths prescribe one or more changes in $x$ to reach $z$. The extreme perturbations in one or more features could lead $z$ to be pointed outside the plausible space (a region not covered by the training set). To overcome the issue of implausibility, we restrict the perturbations to a neighbourhood of $x$ in the desired space (e.g., the loan-approved space in Fig.~\ref{decision-surface-cfs}), adhering to user constraints. This neighbourhood is computed from the desired space using a KD-Tree (a k-dimensional space-partitioning data structure). Some studies use these data instances in the neighbourhood as counterfactuals. However, this way of doing is not encouraged in recent papers due to the concern of data leakage \cite{dexe2020empirical, dataleakageprivacy, ISJ-rajapaksha2020lormika}. We utilise these neighbours for further perturbations to meet the Eq.~\ref{eq-1}. The process of perturbations (described in Sec.~\ref{perturbations}) is guided by the MI shared among the features.  

\subsection{Perturbations}\label{perturbations}
In this section, we describe the process of perturbations in $x$ to generate the counterfactual instance $z$ with the aim of answering to the following questions:
\begin{itemize}
    \item Question (i): how the features to perturb in $x$ are selected?
    \item Question (ii): to what extent are perturbed the selected features?
    \item Question (iii): what mechanism is used to update the feature values?
\end{itemize}

Regarding Question (i), the features are selected from the MI scores and the user-specified list of features to be modified. As our approach upholds perturbations to the subset of features, accordingly, there are three options. The first method attempts to perturb a single feature at a time, exploiting the user-specified list of features. The second and third methods perturb double and triple features simultaneously, respectively, and use tuples formed with MI scores (described in Section~\ref{MI}).

To answer the Question (ii), we can define a Python dictionary (key-value pair, data storage) to store the user preferences regarding the perturbations to each feature, calling it perturbation map as $p =  \{ x_{1}: [p_{1^l}, p_{1^u}],...x_d:[p_{d^l}, p_{d^u}] \}$, where $p_{i^l}$ and $p_{i^u}$ are the lower and upper bound of the user-specified interval for $i^{th}$ feature. For example, if the $i^{th}$ feature represents the `income' of a loan applicant, then $p_{i^l}$ tells by how much the `income' might lower at most, and $p_{i^u}$ tells by how much the `income' might raise at most. Credit managers may be able to define this information precisely from their experience. In addition, lay users can also impose these constraints in agreement with their requirements. For the case of categorical features, as we are dealing only with binary categories, $p_{j^l}$ and $p_{j^u}$ hold the current and new possible value (category) for the $j^{th}$ categorical feature (i.e., if $p_{j^l}$ is $0$, then, $p_{j^u}$ could be $1$).

For answering to Question (iii), we can consider several strategies to identify the relevant and meaningful changes to the input features by following a specific search in the actual space. For example, gradient-based approaches (optimisation) search for perturbations that minimise the difference between $x$ and $z$. Such approaches adjust the feature values in the direction of the gradient iteratively. Unfortunately, when user-defined constraints are in place, this strategy could become less effective because of the uncertainty of the state of convergence and higher costs in terms of changes to incur in CEs \cite{dice}. As an alternative, the rule-based search uses a set of predefined rules to guide the search for perturbations, but this approach requires comprehensive domain knowledge to define the rules \cite{lore}. 

Hybrid search includes several approaches together, and its effectiveness depends on the specific problem and the characteristics of the data.
We have customised hybrid search in mixed-feature perturbations (perturbing numerical and categorical features). It involves predicting the values of features using a supervised ML model trained on the available dataset, except for the feature whose value is being predicted. Thus, the values of features are predicted using their respective prediction models, the regressor for numerical and the classifier for categorical features (for each feature, a separate model is trained to predict its outcomes given the rest of known feature values). The predicted values must adhere to $p$ to be considered as a legit perturbation value.

Finally, let us discuss briefly below the three methods that are provided by UFCE for finding out counterfactuals.

The first method is based on single-feature perturbations, in which numerical and categorical features can be perturbed separately. Particularly for this method, we do not predict new values of features from the learning model; rather, we use the values from the subspace formed by $p_{i^l}$ and $p_{i^u}$. This subspace contains the ordered values uniformly distributed from $p_{i^l}$ to $p_{i^u}$. For iterative perturbations, the new feature values are taken from the mid of the subspace by traversing on it using the notion of binary search. In other words, when a perturbed instance $z$ does not provide the desired outcome, then, to continue the cycle of perturbations, the next value is taken from the centroid of the subspace. In the case of categorical features, the category is reversed (i.e., $0$ to $1$ or $1$ to $0$).

The second and third methods simultaneously perform double and triple feature perturbations. The first feature coming in the tuple follows the same notion of single feature perturbations. After perturbing the first feature, the value of the second feature is predicted from the ML model previously trained. Consequently, both feature values help in the prediction of the value for the third feature. This process is repeated until a valid and plausible $z$ is found (further details are provided in Section~\ref{algo-details}).

\subsection{Algorithmic Details}\label{algo-details}
Algorithm~\ref{alg:ufce} presents the pseudocode for UFCE and its sub-components. The input parameters of UFCE are test instance $x$, perturbation map $p$, $desired\_space$ (e.g., loan approved training space), categorical features $catf$, numerical features $numf$, list of protected features $protectf$ (these are decided with the domain knowledge and their obvious nature such as Family cannot be suggested to increase or decrease), list of all features ($features$), desired outcome $t$, black-box model $f$, the training data $X$, and a dictionary $step$ (holding the feature distribution to be used in single feature method). The output is a set of CEs ($C_{cf}$). The list of features to change $f2change$ are obtained from the keys of $p$ (line 1). The MI among the features is computed by calling the sub-routine CMI (short for COMPUTE\_MUTUAL\_INFORMATION, line 2), which provides a sorted list of feature pairs $mi\_{pair}$ with MI scores. The nearest neighbourhood $nn$ of $x$ is mined by calling a sub-routine FNN (short for FIND\_NEAREST\_ NEIGHBOURS) in a desired space within a specific radius such that $y=t$ (line 3). A neighbourhood (subspace) is computed which adheres to user-constraints in terms of feature ranges. This subspace is created by calling a routine INTERVALS that takes input of $nn$, $p$, $f2change$, $x$; and outputs the subspace that adheres to user-constraints and intersects with the neighbourhood in the desired space (line 4). Then, the different variations of perturbations to find CEs (single, double, and triple feature) are called (lines 5-7), and they return the counterfactual instances $z^1$, $z^2$, and $z^3$, respectively, which represent the counterfactuals specifying the single, double, and triple feature changes to achieve the desired outcome $t$. Finally, the set of $C_{cf}$ contains the counterfactuals from all three variations of feature perturbations (line 8).

\begin{algorithm}
\small
\caption{UFCE}
\label{alg:ufce}
\textbf{Input:} $x$, $p$, $desired\_{space}$, $catf$, $numf$, $protectf$, $features$, $t$, $f$, $X$, $step$. \\
\textbf{Output:}\;  \textrm{set of counterfactual explanations}\;($C_{cf}$).
\begin{algorithmic}[1]
\State $f2change \gets p.keys()$ \Comment{features to change.}
\State $mi\_{pair} \gets \textrm{CMI}(features, X)$ \Comment{sorted list of pairs of features.}
\State $nn \gets \textbf{FNN}(x, desired\_{space}, radius)$  \Comment{using k-dimensional tree.}
\State $subspace \gets \textbf{INTERVALS}(nn, p, f2change, x)$  \Comment{harnessing the neighbourhood with user-specified feature ranges.}
\State $z^{1} \gets \textbf{SINGLE\_F}(x, catf, p, f, t, step)$  \Comment{single feature perturbations.}
\State $z^{2} \gets \textbf{DOUBLE\_F}(X, desired\_space, x,           subspace, \newline mi\_{pair},  catf, numf, features,        
      protectf, f, t)$ 
     \Comment{double feature perturbations.}
\State $z^{3} \gets \textbf{TRIPLE\_F}(X, desired\_space, x, 
     subspace, \newline mi\_{pair},  catf, numf, features, protectf, f, t)$ 
     \Comment{triple feature perturbations.}
\State $C_{cf} \gets C_{cf}.add(z^{1}), C_{cf}.add(z^{2}), C_{cf}.add(z^{3})$
\end{algorithmic}
\end{algorithm}

\begin{algorithm}
\small
\caption{UFCE sub-routines}
\label{alg:ufce-subroutines}
\begin{algorithmic}[1]
    \Function{FNN}{$desired\_space, x, radius$}:
        \State $tree \gets KDTree(desired\_space)$
        \Comment{tree holds the k-dimensional space partitions.}
        \State $idx \gets tree.query\_ball\_point(x, r=radius)$ \Comment{idx hold the nearest neighbor indices in a radius.}
        \State $nn \gets tree.data[idx]$ \Comment{nn are the nearest neighbors found on the indices (idx) of tree.data.}
    \State \Return $nn$
    \EndFunction
    \Function{CMI}{$features, X$}:
        \State $feature\_pairs \gets dict()$,
         $mi\_pairs \gets []$ \Comment{be the empty key-value data storage, and empty list of feature pairs.}
        \For{\textrm{each $\qdist{f_i, f_j}$ in $features$}}
            \State $mi\_scores \gets mi\_classif(X[f_i], X[f_j])$
            \State $score \gets mi\_scores[0]$ 
            \State $feature\_pairs[score] \gets \qdist{f_i, f_j}$
        \EndFor
        \State $P \gets dict(sorted(feature\_pairs.items())) $
        \For{\textrm{each $ i $ in $P.keys()$}}
            \State $pair \gets P[i]$
            \If{$pair$ \textrm{not in} $mi\_pairs$} $mi\_pairs.append(pair)$
            \EndIf
        \EndFor
    \State \Return $mi\_pairs$
    \EndFunction
    \Function{INTERVALS}{$nn, p, f2change, x$}:
        \State $subspace \gets dict()$
        \Comment{subspace be the empty key-value pair data storage for each feature.}
        \For{\textrm{each $ i $ in $p$}}
            \State $lower \gets p[i][0] $, $upper \gets p[i][1] $ 
            \If{$upper \ge nn[i].max()$}
                 $subspace[i] \gets [lower, nn[i].max()]$
            \ElsIf {$lower \le nn[i].min()$}
                 \State $subspace[i] \gets [nn[i].min(), upper]$
            \Else
                \; $subspace[i] \gets [lower, upper]$
            \EndIf
        \EndFor
    \State \Return $subspace$
    \EndFunction
\end{algorithmic}
\end{algorithm}

The Algorithm~\ref{alg:ufce-subroutines} presents pseudocode for the sub-routines (functions) as follows.
\begin{itemize}
    \item The FNN function (lines 1-5) has three arguments, namely $desired\_space$, $x$, and $radius$. It creates a KDTree object, which holds the k-dimensional space partitions based on the $desired\_space$. Then, it finds the indices of the nearest neighbours of $x$ within a certain $radius$ using the $query\_ball\_point$ method. Finally, it returns the nearest neighbours found using the indices.
    \item The CMI function (lines 6-16) has two arguments, namely $features$ and $X$. It initialises an empty dictionary $feature\_pairs$ (key-value storage) and an empty list $mi\_pairs$. It then iterates through all the feature pairs in $features$ and computes their MI scores using the $mi\_classif$ (short for $mutual\_info\_classif$) function provided by scikit-learn (line 9), for conciseness, we represent feature pair with $\qdist{f_i, f_j}$, the actual implementation follows the structure of nested for-loop). The MI score is used as a key in $feature\_pairs$ to store the corresponding feature pair. The $feature\_pairs$ dictionary is sorted in descending order of the MI scores and stored in a new dictionary $P$ (storing feature pairs). Finally, the function iterates through the keys of $P$ and appends the corresponding feature pair to $mi\_pairs$ if it does not already exist. The function returns $mi\_pairs$.
    \item The INTERVALS function (lines 17-25) has four arguments, namely $nn$, $p$, $f2change$, and $x$. It initialises empty storage $subspace$ (key-value). It then iterates through all the features in $p$ (perturbation map) and sets their corresponding lower and upper bounds. If the upper bound is greater than or equal to the maximum value of the corresponding nearest neighbourhood, then the upper bound is set to the maximum value of the neighbourhood. Similarly, the lower bound is validated (it is verified in agreement with user constraints because it could be large enough to fall outside the actual distribution). The lower and the upper bounds are then stored in the $subspace$ using the feature as the key. The function returns the $subspace$.
\end{itemize}

\begin{algorithm}[hbt!]
\caption{UFCE (Single\_F)}
\label{alg:ufce-sf}
\small
\begin{algorithmic}[1]
    \Function{Single\_F}{$x, catf, p, f, t, step$}:
        \For{\textrm{each $ i $ in $p$}}
            \State $start \gets p[i][0]$, $end \gets p[i][1]$ 
            \If{$i$ \textrm{not in} $catf$}
                \While{$ start <= end $}
                    \State $tempdf \gets x$, $mid \gets start+(end-start)/2$
                    \State $tempdf.loc[:,i] \gets mid $
                    \If{$f(tempdf) = t$ \textrm{AND tempdf is plausible}}
                        \State $z \gets tempdf$
                        \State $end \gets mid - step[i]$
                    \Else \; \State $start \gets mid + step[i]$
                    \EndIf
                \EndWhile
            \Else
            \State $z \gets x$ , $z.loc[:,i] \gets (1-end) $
                \If{$f(z) = t$ \textrm{AND z is plausible}} \Return $z$
                \EndIf
            \EndIf
        \EndFor
    \State \Return $z$
    \EndFunction
    \end{algorithmic}
\end{algorithm}

The Algorithm \ref{alg:ufce-sf} presents pseudocode for single feature perturbations of UFCE as follows.
The function Single\_F (lines 1-16) has the following input parameters: $x$, $catf$, $p$, $f$, $t$, and $step$. The input instance $x$ is to be explained.
The function iterates over for each feature $i$ in the feature map $p$. Suppose $i$ is not a categorical feature. In that case, the function performs a binary search-inspired traversing on the feature values to find the minimum value $mid$ such that changing the $i^{th}$ feature value of $x$ to $mid$ will result in the target outcome $t$ and a plausible explanation $z$ (plausibility is verified by using the outlier detection algorithm, LOF, described in Sec.~\ref{cf-properties}). If the binary search fails to find such a value in the range $[start, end]$, where $start$ and $end$ are the lower and the upper bounds of the feature range (this subspace is discretised uniformly), the search goes on in the lower and upper half from the mid, where $step$ is a dictionary holding the step size for each feature used to traverse to the next element.
If $i$ is a categorical feature, the function sets the feature value of $i^{th}$ feature in $z$ to its reverse value $1-end$ and checks if $f(z) = t$ and $z$ is a plausible explanation. If the condition is met, $z$ is returned. Finally, the function returns the resulting explanation $z$.

\begin{algorithm}[hbt!]
\caption{UFCE (Double\_F)}
\label{alg:ufce-df}
\small
\begin{algorithmic}[1]
    \Function{Double\_F}{$X, x, subspace, mi\_pair, catf, \newline numf, features, protectf, f, t$}:
        \For{\textbf{each} $pair$ \textrm{in} $mi\_pair$}
        \State $i \gets pair[0]$, $j \gets pair[1]$, $z \gets x$ \Comment{instantiate z with x.}
        \If{$i$ \textrm{and} $j$ \textrm{in} $subspace$}
            \If{($i$ \textrm{in} $numf$ \textrm{and} ($j$ \textrm{in} $numf$ \textrm{or in} $catf$))  and ($i$ \textrm{and} $j$ \textrm{not in} $protectf$) }
                \State $start_1 \gets subspace[i][0]$
                \State $end_1 \gets subspace[i][1]$ 
                \State $h \gets \textrm{regress\_model}(X, i, j)$
                \State $g \gets \textrm{classif\_model}(X, i, j)$ 
                \State $traverse\_space \gets sorted(random.uniform(start_1, end_1))$
                \While{$\textrm{length of traverse\_space is not empty}$}
                    \State $mid \gets start_1+(end_1-start_1)/2$ \State $z.loc[:,i] \gets traverse\_space[mid]$
                    \State $z = z.loc[:, z.columns\;!= j]$
                        \If{$j$ \textrm{in} $numf$}
                          $new\_j = h(z)$
                          $z.loc[:,j] \gets new\_j$
                        \Else \; $new\_j = g(z)$, $z.loc[:,j] \gets new\_j$
                        \EndIf
                    \If{$f(z) = t$ \textrm{AND z is plausible}} \Return $z$
                    \Else \; \textrm{try: delete} $traverse\_space[:mid]$ \textrm{except pass} 
                    \EndIf
                \EndWhile
            \EndIf
            \hspace{8pt} \vdots
        \If{($i$ \textrm{and} $j$ \textrm{in} $catf$) and ($i$ \textrm{and} $j$ \textrm{not in} $protectf$)}
                \State $z.loc[:,i] \gets subspace[i][1]$
                \State $z.loc[:,j] \gets subspace[j][1]$
                \If{$f(z) = t$ \textrm{AND z is plausible}} \Return $z$
                \EndIf
            \EndIf
        \EndIf
        \EndFor
    \State \Return $z$
    \EndFunction
\end{algorithmic}
\end{algorithm}

The Algorithm \ref{alg:ufce-df} presents pseudocode for double feature perturbations of UFCE as follows.
The function $Double\_F$ (lines 1-23) has the following input arguments: $X$, $x$, $subspace$, $mi\_pair$, $catf$, $numf$, $features$, $protectf$, $f$, and $t$. This function aims to search for a data point $z$ that satisfies the condition $f(z) = t$ while performing double-loop perturbations on the input $x$.

The function first iterates over each pair in $mi\_{pair}$, a list of pairs of the features ordered with their MI scores. It then checks whether $i$ and $j$ features are in the valid subspace. If so, then, it checks whether $i$ is in $numf$ and $j$ is in $numf$ or $catf$ and if they do not belong to $protectf$ (protected features). If all these conditions are satisfied, then it generates a uniform random set of values within the range ($start$, $end$) of $i^{th}$ feature as $traverse\_space$. It iterates over each value in $traverse\_space$, the sorted set of uniform random values. In the meantime, a regressor $h$ and classifier $g$ are trained to predict the feature $j$ value. To predict the $j$ value, they use $z$ that contains the copy of $x$ and sets the value of $i^{th}$ feature to the $mid$ value of $traverse\_space$. The function then removes the feature (column) corresponding to $j^{th}$ feature from $z$, and depending on whether $j$ is a categorical or numeric feature, it applies $h$ or $g$ to predict the new value for $j^{th}$ feature. It checks whether the resulting data point satisfies the condition $f(z) = t$ and, if so, returns it. Otherwise, it reduces the size of the $traverse\_space$ (deleting the values from start to midpoint) and continues to the next value. In the prediction mechanism, the first feature provides a space to move for more perturbations (in the uniform distribution or respective feature distribution from start to end). It predicts the second feature value from the respective predictor. After line $18$, the vertical dotted line represents more cases (if possible) of different combinations of numeric and categorical features (handled accordingly).

If $i$ and $j$ are both categorical features and are not present in $protectf$, the $Double\_F$ function sets $i$ and $j$ features to the maximum value within their respective $p$ in subspace (reverse of values) and checks whether the resulting data point $z$ satisfies the condition $f(z) = t$. If so, it returns $z$.

\begin{algorithm}[t!]
\caption{Computing distance of Counterfactual Explanations using $\delta$}
\label{alg:bestcfs}
\textbf{Input:} $x$, $Z$, $t$, $f$, $\lambda$ ; \
\textbf{Output:} suitable counterfactual explanation $z^*$.
\begin{algorithmic}[1]
\State $z^* \gets \textrm{initialise with copy of $z\in Z$}$
\State $\delta^* \gets +\infty$
\For {each $z \in Z$}
\State $\delta(z,x) \gets  prox\_Jac(z,x) + \lambda \cdot prox\_Euc(z,x)$
\If{$\delta(z,x) < \delta^*$}
\State $z^* \gets z$
\State $\delta^* \gets \delta(z,x)$
\EndIf
\EndFor
\State \textbf{return} $z^*$
\end{algorithmic}
\end{algorithm}

The Algorithm~\ref{alg:bestcfs} finds the most suitable CE $z^*$ that satisfies the desired outcome $t$ by iteratively computing the distance between the given test instance $x$ and all the possible counterfactual instances $z \in Z$ that satisfy the outcome $f(z) = t$ (i.e., candidate counterfactuals). The distance metric $\delta(z, x)$ is computed by a weighted addition of $prox\_Jac$ and $prox\_Euc$ where the former measures the distance between categorical features, and the later measures the Euclidean distance between numerical features. The algorithm returns the instance $z^*$ with the smallest distance $\delta^*$.

\section{Experiments and Results}\label{experiments}
We concentrate on empirical findings that help us respond to what we see as essential research questions:
\begin{itemize}
    \item \textbf{(RQ1)} Does user feedback (user constraints) affect the quality and computations of CEs? 
    \item \textbf{(RQ2)} How do randomly taken user constraints affect the generation of CEs? 
    \item \textbf{(RQ3)} What is the behaviour of UFCE on multiple datasets? 
\end{itemize}
We have performed three experiments to answer the three research questions. Experimental settings are presented in Section~\ref{exp-settings}. Then, the next Sections (\ref{sec:rq1}, \ref{sec:rq2}, and \ref{sec:rq3}) answer to the above research questions. 

\subsection{Experimental Settings} \label{exp-settings}

\begin{table*}[t]
\centering
\caption{Dataset Details: Size (\#instances), Features (total features), Num-Cat (\#Numerical and \#Categorical features), Classes (total classes), Positive Class (Percentage), and average 5-fold cross-validation (CV) accuracy with standard deviation for Logistic Regression.}
\label{tab:datasets}
\resizebox{\textwidth}{!}{%
\begin{tabular}{lllllllll}
\toprule
dataset     & Size & Features & Num|Cat & Classes & Positive Class & LR (CV Acc.) \\ \hline 
Graduate     & 500     & 7          & 6 | 1     & 2       & Chance of Admit ($81\%$) & $0.86 \pm 0.04$ \\ 
Bank Loan     & 7636     & 11     & 5 | 4     & 2       & Loan granted ($41\%$)   & $0.97 \pm 0.06$ \\ 
Red-Wine      & 1598     & 11      & 11 | 0      &  2        & Good quality wine ($53\%$) & $0.74 \pm 0.04$ \\ 
Bupa     & 345    & 6   & 6 | 0     & 2       & No Alcoholism ($57\%$)   & $0.73 \pm 0.04$ \\ 
Movie    & 505      & 18          & 17 | 1    & 2       & Nominated for Oscar($54\%$) & $0.68 \pm 0.04$ \\ 
\bottomrule
\hline
\end{tabular}%
}
\end{table*}

\subsubsection{Data sets}\label{datasets}
We utilised five datasets to test the CE methods under study. We choose two datasets with mixed data types from Kaggle competitions\footnote{\url{https://www.kaggle.com/}} publicly available and three datasets from the KEEL-Dataset-repository\footnote{\url{https://sci2s.ugr.es/keel/datasets.php}} \cite{keel-datarepository}, to provide readers with complete data analysis. All these datasets are for binary classification and they can be downloaded from the UFCE project repository\footnote{\url{https://github.com/msnizami/UFCE/tree/main/data}} in the format required to run the experiments to be described in the rest of this paper. Detailed information on the datasets (i.e., their name, size, number of features, numerical and categorical counts, number of classes, and percentage of the positive class) is presented in Table~\ref{tab:datasets}.

\subsubsection{Machine Learning Model}
\label{mlmodels}
We choose Logistic Regression (LR) as the classification model with default hyper-parameters. We trained LR with the same hyper-parameters for all explainers to establish consistency. The performance measures for LR are presented in terms of average (avg.) accuracy of 5-fold CV for all datasets in Table~\ref{tab:datasets}.

\subsubsection{Counterfactual Explainer Methods}
\label{CE-methods}
DiCE \cite{mothilal2020explaining} strives to provide \textit{diverse} CEs; it provides an implementation that also covers categorical features. In our experiments, we implemented and used the standard DiCE\footnote{\url{https://github.com/interpretml/DiCE}} library for results comparison.
The AR \cite{ar} focuses on the issue of actionability. It also adheres to restrictions that prevent immutable features from being altered. The AR explainer works with LR, and we implemented it using the \textit{actionable-recourse}\footnote{\url{https://github.com/ustunb/actionable-recourse}} library. The reason to choose DiCE and AR is due to their support available for imposing user constraints in their respective public libraries. The proposed approach UFCE is model and data-agnostic for tabular datasets.

\subsection{(RQ1) Effects of User-constraints on the Performance and Computation of Counterfactual Generations}\label{sec:rq1}
\begin{table*}[hbt!]
\caption{(RQ1) The performance comparison in terms of generation of feasible counterfactuals ($\%$) for very limited (VL), limited (L), medium (M), flexible (F), and more flexible (MF) constraints. Plaus refers to number of plausible CEs, Act to number of actionable CEs, and Feas to number of feasible CEs.}
\label{tab:perfpercent}
\resizebox{\textwidth}{!}{%
\fontsize{8.75pt}{9.75pt}\selectfont
\begin{tabular}{|l|lll|lll|lll|lll|lll|lll|}
\toprule
- & \multicolumn{3}{c}{UFCE-1}&\multicolumn{3}{c}{UFCE-2} &\multicolumn{3}{c}{UFCE-3} & \multicolumn{3}{c}{AR}&\multicolumn{3}{c}{DiCE}&\multicolumn{3}{c}{DiCE-UF} \\ \midrule
Levels & Plaus & Act & Feas & Plaus & Act & Feas & Plaus & Act & Feas & Plaus & Act & Feas& Plaus & Act & Feas& Plaus & Act & Feas  \\ \hline
VL & 8 & 15  & 8 (16\%)  & 18 & 18 & 18 (36\%)  & 30  &30  & 30 (60\%)  &49   &3  & 3 (6\%)  &40  &4  & 4 (8\%) & 15 & 25 & 15 (30\%)    \\
L  & 9 & 15 & 9 (18\%)  &21  &21  & 21 (42\%)  &37  &37  & 37 (74\%)  &49   &5  & 5 (10\%)  &41  &5  & 5 (10\%) &19 &32 &19 (38\%)      \\ 
M & 12 & 14 & 12 (24\%) & 29 & 29 & 29 (58\%) & 44 & 44 & 44 (88\%)  &49   & 5& 5 (10\%) & 43 &6  & 6 (12\%) &20 &33 &20 (40\%)      \\ 
F  & 17 & 21 & 17 (34\%) &37  &37  & 37 (74\%)  &47  &47  & 47 (94\%)  &49   &10 & 10 (20\%) & 46 &9  &9 (18\%) &26 &39 &26 (52\%)      \\ 
MF  &19  &23  & 19 (38\%)  &44  &44  & 44 (88\%)  &48  & 48 & 48 (96\%) &49   &8  &8 (16\%)  &38  &8  &8 (16\%)  &28  &42  &28 (56\%)  \\  \hline
Average  &-  &-  &  26.0\%  &-  &-  &  59.6\%  &-  & - &  82.4\% &-   &-  &  12.4\%  &-  &-  &  12.8\%  &-  &-  &    43.2\%    \\  \bottomrule
\end{tabular}
}
\end{table*}
\begin{figure*}[hbt!]
    \includegraphics[width=.24\textwidth]{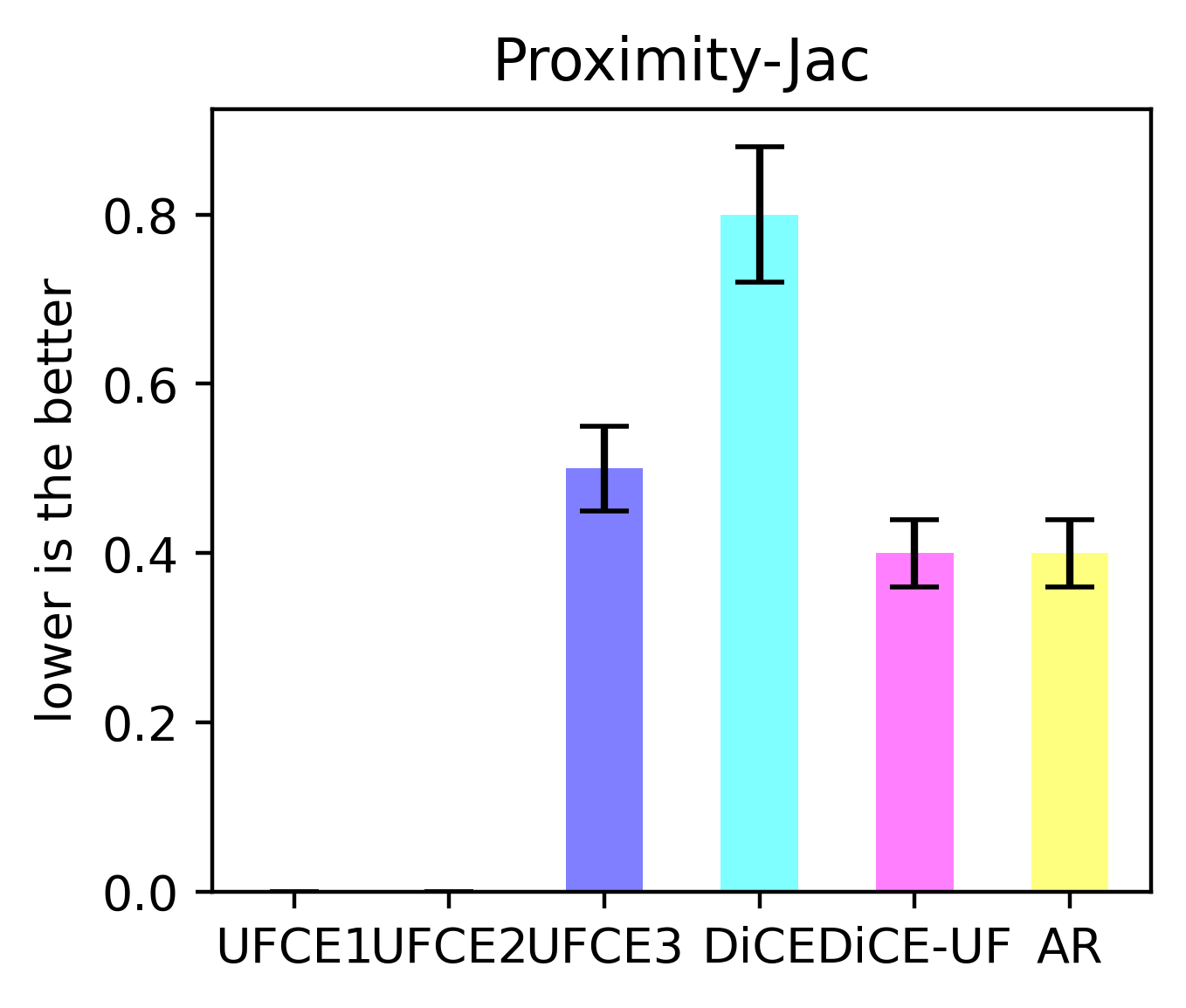}\hfill
    \includegraphics[width=.24\textwidth]{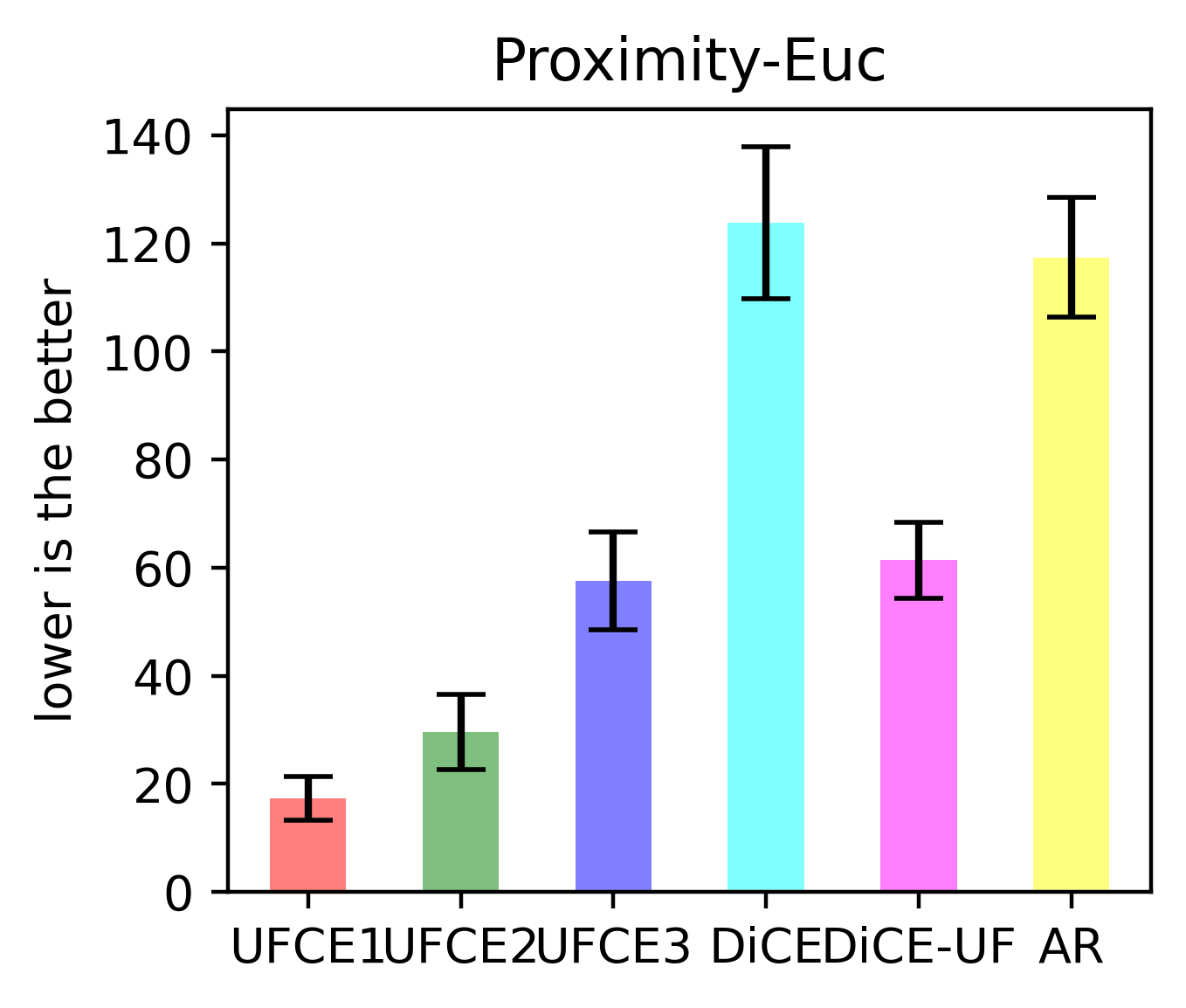}\hfill
    \includegraphics[width=.24\textwidth]{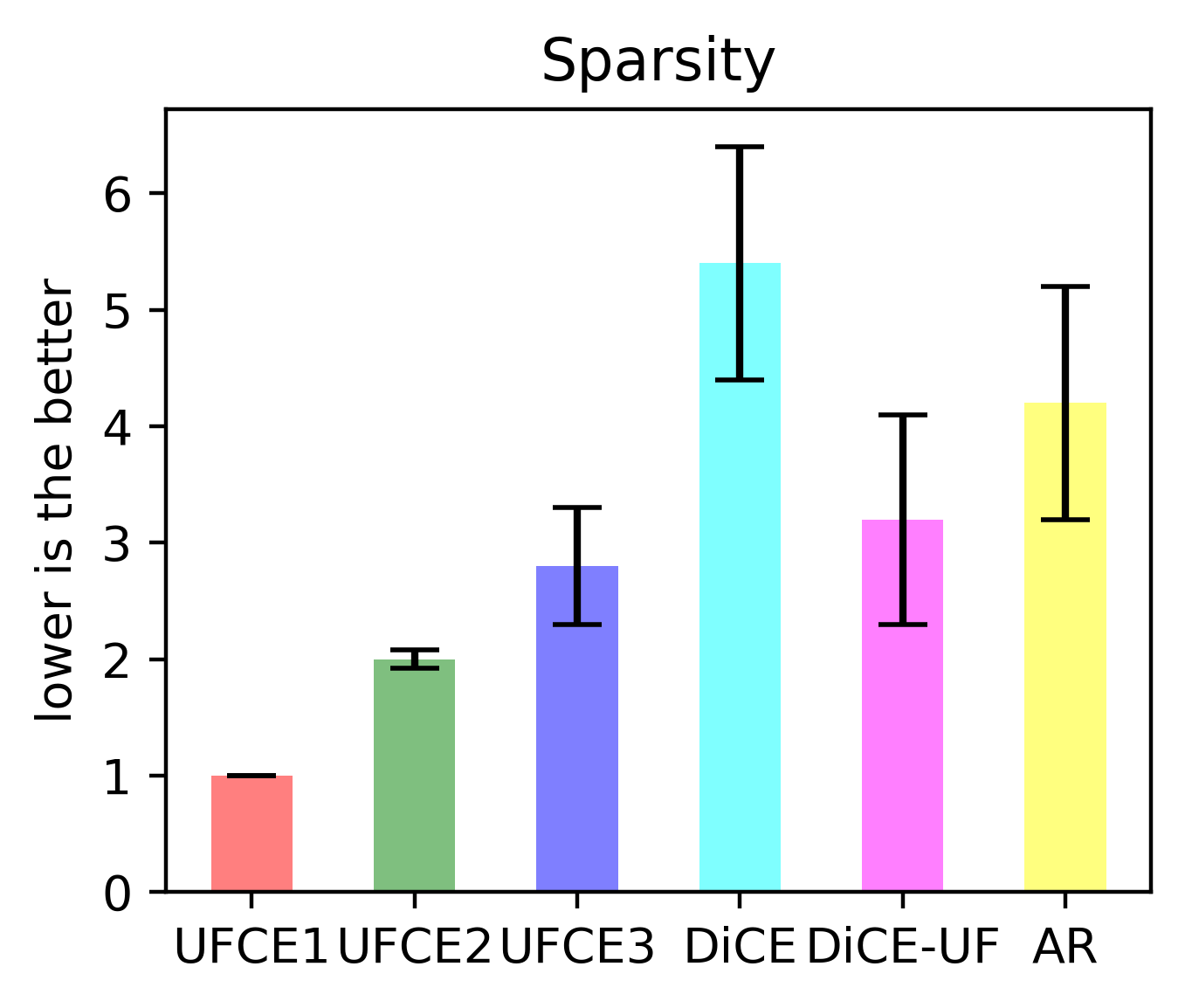}\hfill
    \includegraphics[width=.24\textwidth]{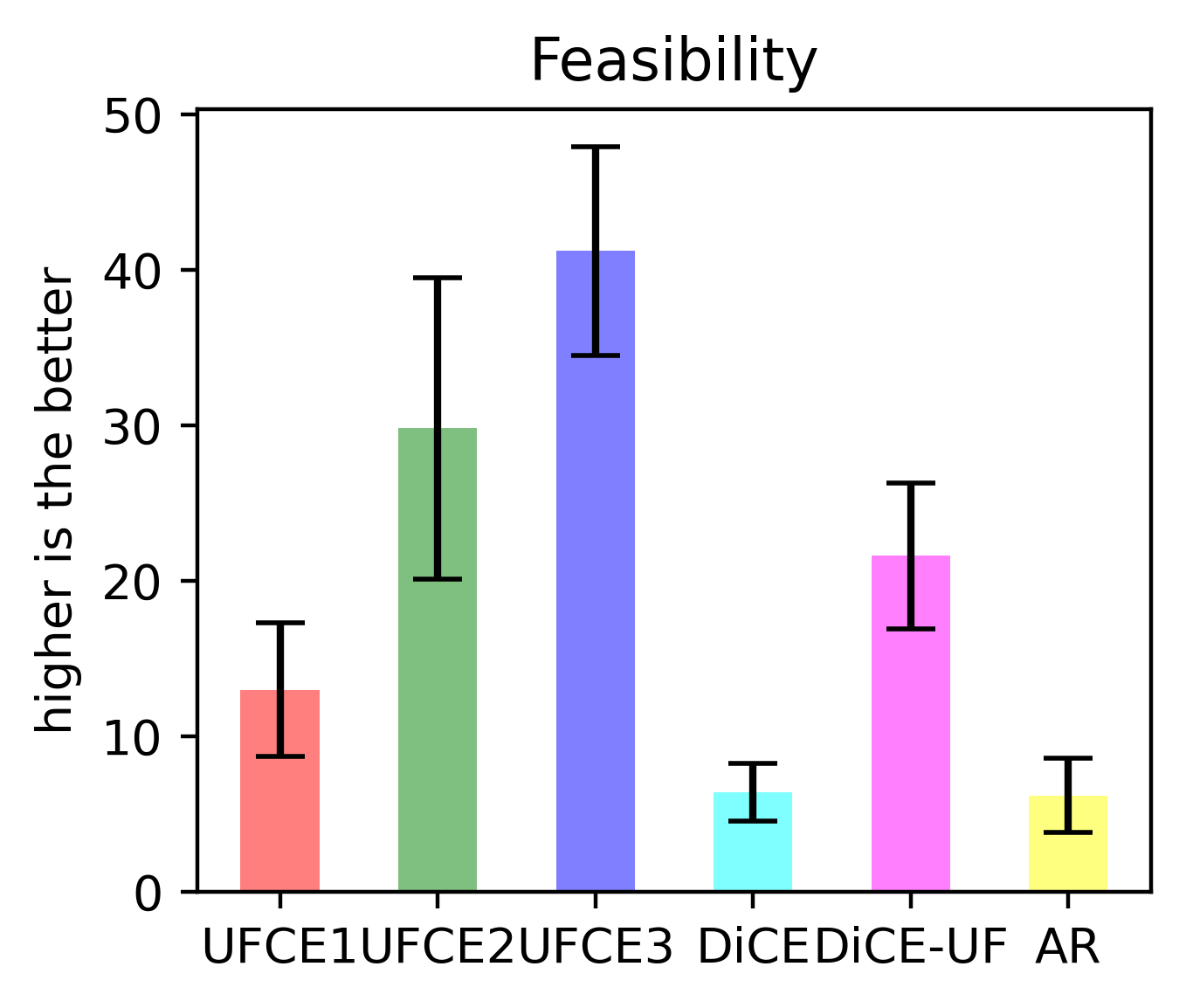}
\caption{(RQ1) Performance of CE methods for different evaluation metrics (with error bar of st.dev).}
\label{fig:exp1}
\end{figure*}
This experiment entails the details of how the different levels of user constraints (user feedback) can affect the performance of the generation of CEs. The different levels of user constraints are configured to perturb the test instances to generate CEs.
These constraints help to form the perturbation map $p$ that guides the sub-processes of UFCE to generate counterfactuals. 
A specific percentage (absolute value) of median absolute deviation from the actual data distribution is computed as a user-specified perturbation limit for each numeric feature. These configurations are divided into five levels and termed as \textit{very limited, limited, medium, flexible, and more flexible}. These levels are assumed to simulate the scenarios when different users can specify different choices. The different levels of choices simulate the behaviour of a user in the real scenario as follows:
\begin{itemize}
    \item Very limited - This value is a $20\%$ of the median absolute deviation of the relevant data.
    \item Limited - This value is a $40\%$ of the median absolute deviation of the relevant data.
    \item Medium - This value is a $60\%$ of the median absolute deviation of the relevant data.
    \item Flexible - This value is a $80\%$ of the median absolute deviation of the relevant data.
    \item More flexible - This value is a $100\%$ of the median absolute deviation of the relevant data.
\end{itemize}
The Bank Loan dataset is considered for this experiment.
For example, the median absolute deviation of the feature `Income' is $50.10$. Accordingly, in this case, `very limited' corresponds to $10.02$, `limited' is $20.04$, `medium' is $30.06$, `flexible' is $40.08$, and `more flexible' is $50.10$. 

The lower bound $p_{i^l}$ of the perturbation map $p$ is initialised by copying the $x_i$ value and the upper bound $p_{i^u}$ with a value by adding the respective percentage (i.e., $20\%$, $40\%$, $60\%$, $80\%$, and $100\%$) in $x_i$ for the $i^{th}$ feature. 
This process is repeated for all the features taking part in perturbations. For categorical features, the feature values are reversed in all five levels of constraints.
The $p$ is updated iteratively for each level of user constraints, and the respective counterfactuals are computed. 

We run the experiment on a pool of $50$ test instances, for each test instance the counterfactuals are generated for all levels of user constraints with UFCE, DiCE, and AR (our approach includes its $3$ variations). The DiCE was configured in two ways: (i) DiCE-UF takes as input the same user feedback as UFCE; and (ii) the basic DiCE does not take as input any specific user feedback but after counterfactuals are generated we verify if they adhere or not to the desired user feedback ranges. The AR was configured with input features which were suppose to be changed according to user feedback, and its generated counterfactuals were checked afterwards whether they adhere to user feedback or not. All the features are assumed as the user-specified list of features to change for all methods. 


For each test instance, each counterfactual explainer was configured to give a chance to generate its $5$ best counterfactuals. Then, costs of proximity were calculated for each CE, and the one nearest to test instance was chosen (given that it is feasible) to consider for further evaluations. A CE is feasible if it is actionable and plausible. To fulfill this requirement, we had considered a CE as an actionable when it used at least $30\%$ of the features from the user-specified list to its total changes (suggested feature changes) and it is not an outlier. Table~\ref{tab:perfpercent} presents the consolidated results of feasible counterfactuals ($\%ge$) by each method for all five levels of user
feedback, and Fig.~\ref{fig:exp1} plots the average results for all evaluation metrics. Similarly, the time is noted per counterfactual (in seconds) for all methods and presented in Table~\ref{tab:time}.

\begin{table}[hbt!]
\centering
\caption{The performance of different CE methods, in terms of average time per CE (seconds).}\label{tab:time}
\fontsize{8.75pt}{8.75pt}\selectfont
\begin{tabular}{|l|l|l|l|l|l|l|}
\toprule
- & UFCE1 & UFCE2 & UFCE3 & AR & DiCE & DiCE-UF \\ \midrule
Time & 0.28 & 0.47 & 1.10 & 0.41 & 0.51 & 0.61 \\ \bottomrule
\end{tabular}
\end{table}
The performance of generating feasible counterfactuals gets better as we move from `very limited' to `more flexible' in Table~\ref{tab:perfpercent}. In general, UFCE2 and UFCE3 performed better than the other methods. UFCE3 took more time on an average than other methods, when `more flexible' user constraints are in place, it takes more time. The reason behind higher time for UFCE3 is due to multiple combination of features and wider subspace to explore.


In general, UFCE surpassed DiCE, DiCE-UF, and AR in all configurations of user constraints for a feasible counterfactual generation. Regarding computational time, UFCE1 and AR were faster than the other methods to generate counterfactuals. The reasons behind the better performance of UFCE in general are the targeted perturbations to look for valid counterfactuals, plausible to the reference population and actionable to certain user-defined limits.   

Further, in Fig.~\ref{fig:exp1}, the average results for different evaluation metrics are plotted. For each plot, the CE methods are placed on the x-axes and the metric scores on the y-axes. The lower value is the better case for $Proximity-Jac$, $Proximity-Euc$, and $Sparsity$, while the higher value is the better case for $Feasibility$. Proximity-Jac represents the percentage of categorical features utilised. UFCE1 and UFCE2 did not consider any categorical features for generating CEs, and DiCE is the method  utilising maximum categorical features for CEs. Proximity-Euc represents to Euclidean distance of generated CE from the test instance, DiCE turn out to be the most expensive method to suggest changes, whereas UFCE variations performed better than the other methods. Sparsity represents to the number of features changed in the generated CE. DiCE has shown a higher sparsity value, therefore, it incurred multiple feature changes, leading to higher Proximity-Euc, while UFCE performed better, in general. Similarly, UFCE performed better in generating feasible counterfactuals than the other methods.  

This experiment has shown that the impact of user feedback on the generation of counterfactuals is influencing. It is evident that as the user constraints are flexible (at least equal to the median absolute deviation), the results are better for each method incorporating user feedback, in their capacity.  

\subsection{(RQ2) How do the randomly taken user-preferences affect the generation of CEs?}\label{sec:rq2} 
\begin{table*}[hbt!]
\caption{(RQ2) The performance comparison in terms of generation of feasible CE (in $\%ge$) for Monte Carlo-like random generation of user feedback. Plaus refers to number of plausible CEs, Act to number of actionable CEs, and Feas to number of feasible CEs.}
\label{tab:stresstest}
\resizebox{\textwidth}{!}{%
\fontsize{7.75pt}{8.75pt}\selectfont
\begin{tabular}{|l|lll|lll|lll|lll|lll|lll|}
\toprule
- & \multicolumn{3}{c}{UFCE-1}&\multicolumn{3}{c}{UFCE-2} &\multicolumn{3}{c}{UFCE-3} & \multicolumn{3}{c}{AR}&\multicolumn{3}{c}{DiCE}&\multicolumn{3}{c}{DiCE-UF} \\ \midrule
{No.} &  Plaus & Act &  Feas &  Plaus &  Act &  Feas &  Plaus &  Act &  Feas &  Plaus &  Act &  Feas &  Plaus &  Act &  Feas &  Plaus &  Act &  Feas \\ \midrule
\midrule
1 &           9 &        14 &          9 (18\%) &          43 &        43 &         43 (86\%) &          47 &     47 &         47 (94\%) &        49 &       5 &        5 (10\%) &          37 &         1 &          1 (2\%) &             25 &           38 &            25 (50\%) \\
2 &           11 &        17 &          11 (22\%) &          40 &        40 &         40 (80\%) &          44 &        44 &         44 (88\%) &        49 &       1 &        1 (2\%) &          41 &         2 &          2 (4\%) &             19 &           26 &            19 (38\%) \\
3 &           10 &        13 &          10 (20\%) &          39 &        39 &         39 (78\%) &          47 &        47 &         47 (94\%) &        49 &       4 &        4 (8\%) &          41 &         3 &          3 (6\%) &             21 &           34 &            21 (42\%) \\
4 &           15 &        19 &          15 (30\%) &          42 &        42 &         42 (84\%) &          47 &        47 &         47 (94\%) &        49 &       4 &        4 (8\%) &          42 &         2 &          2 (4\%) &             20 &           32 &            20 (40\%) \\
5 &           12 &        15 &          12 (24\%) &          31 &        31 &         31 (62\%) &          41 &        41 &         41 (82\%) &        49 &       4 &        4 (8\%) &          44 &         1 &          1 (2\%) &             18 &           29 &            18 (36\%) \\
6 &           16 &        21 &          16 (32\%) &          31 &        31 &         31 (62\%) &          41 &        41 &         41 (82\%) &        49 &       1 &        1 (2\%) &          39 &         2 &          2 (4\%) &             15 &           23 &            15 (30\%) \\
7 &           13 &        17 &          13 (26\%) &          33 &        33 &         33 (66\%) &          41 &        41 &         41 (82\%) &        49 &       3 &        3 (6\%) &          45 &         2 &          2 (4\%) &             17 &           23 &            17 (34\%) \\
8 &           9 &        15 &          9 (18\%) &          43 &        43 &         43 (86\%) &          45 &        45 &         45 (90\%) &        49 &       4 &        4 (8\%) &          40 &         3 &          3 (6\%) &             23 &           32 &            23 (46\%) \\
9 &           13 &        18 &          13 (26\%) &          47 &        47 &         47 (94\%) &          48 &        48 &         48 (96\% &        49 &       6 &        6 (12\%) &          43 &         2 &          2 (4\%) &             24 &           39 &            24 (48\%) \\
10 &           12 &        14 &          12 (24\%) &          30 &        30 &         30 (60\%) &          40 &        40 &         40 (80\%) &        49 &       1 &        1 (2\%) &          42 &         3 &          3 (6\%) &             14 &           23 &            14 (28\%) \\ \midrule
Avg. &           - &        - &          24\% &          - &        - &         75.8\% &          - &       - &         88.2\%)&        - &       - &        6.6\% &          - &         - &          4.2\%&             - &           - &            39.2\% \\ \bottomrule
\end{tabular}%
}
\end{table*} 
\begin{figure*}[hbt!]
    \includegraphics[width=.24\textwidth]{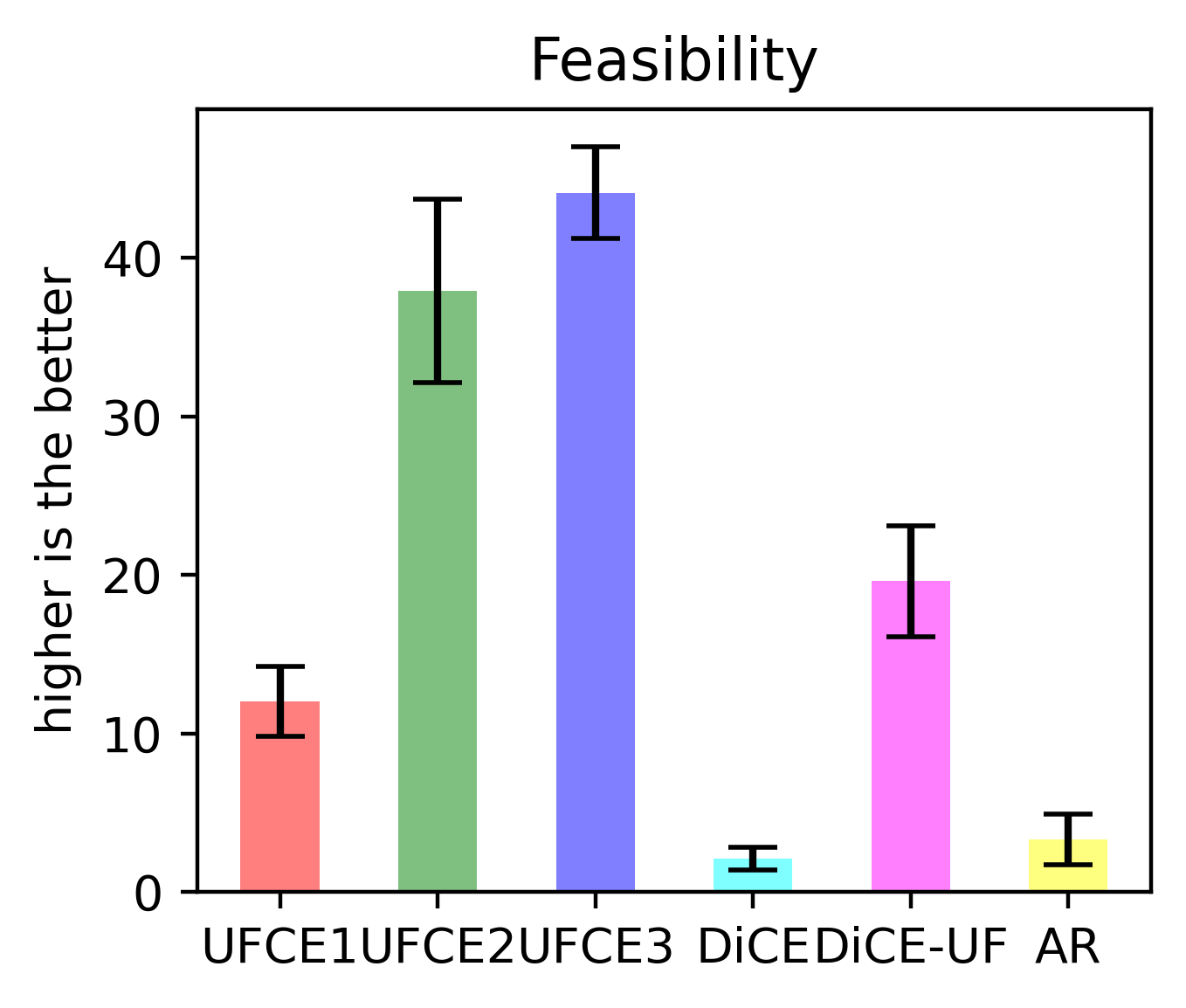}\hfill
    \includegraphics[width=.24\textwidth]{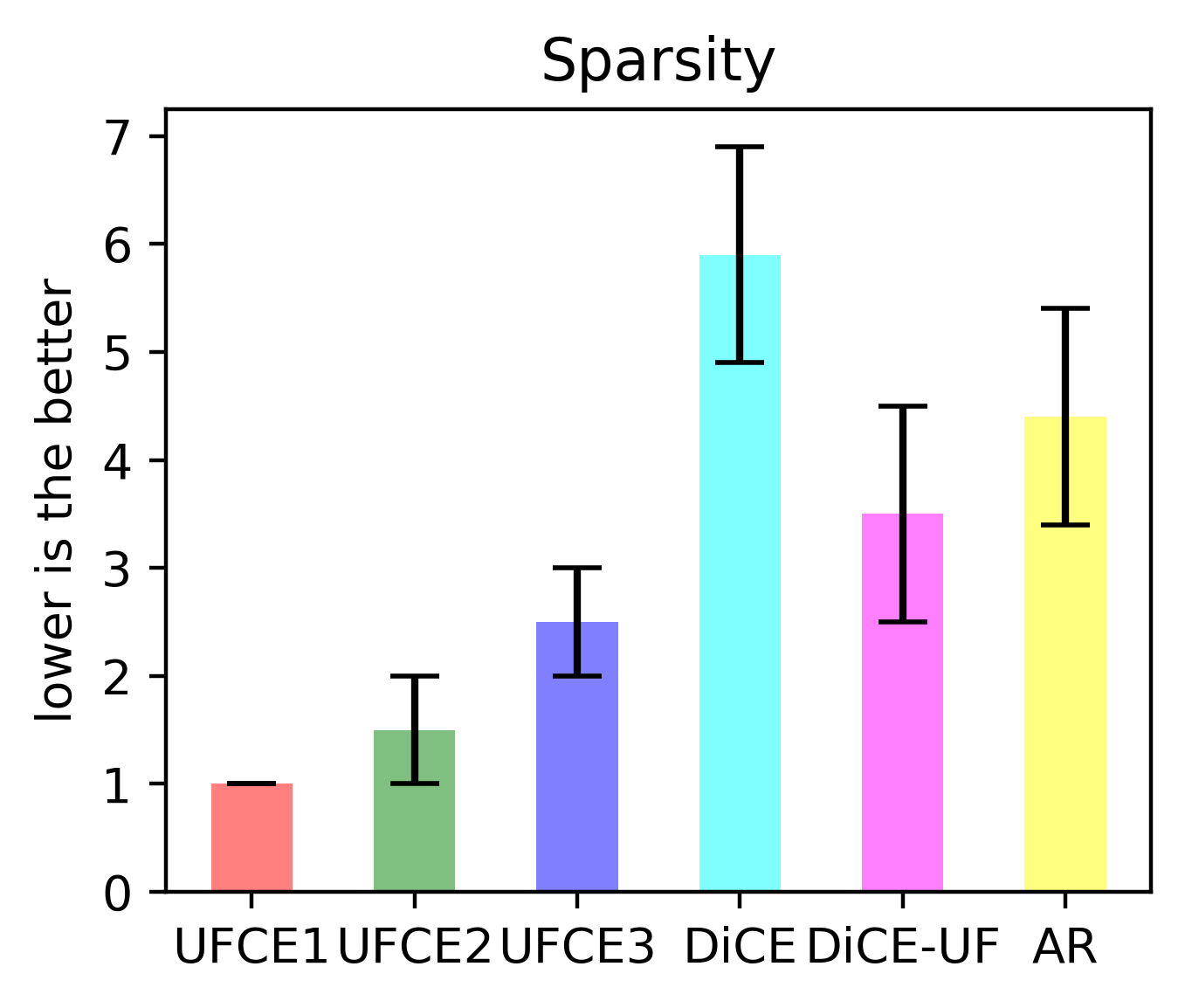}\hfill
    \includegraphics[width=.24\textwidth]{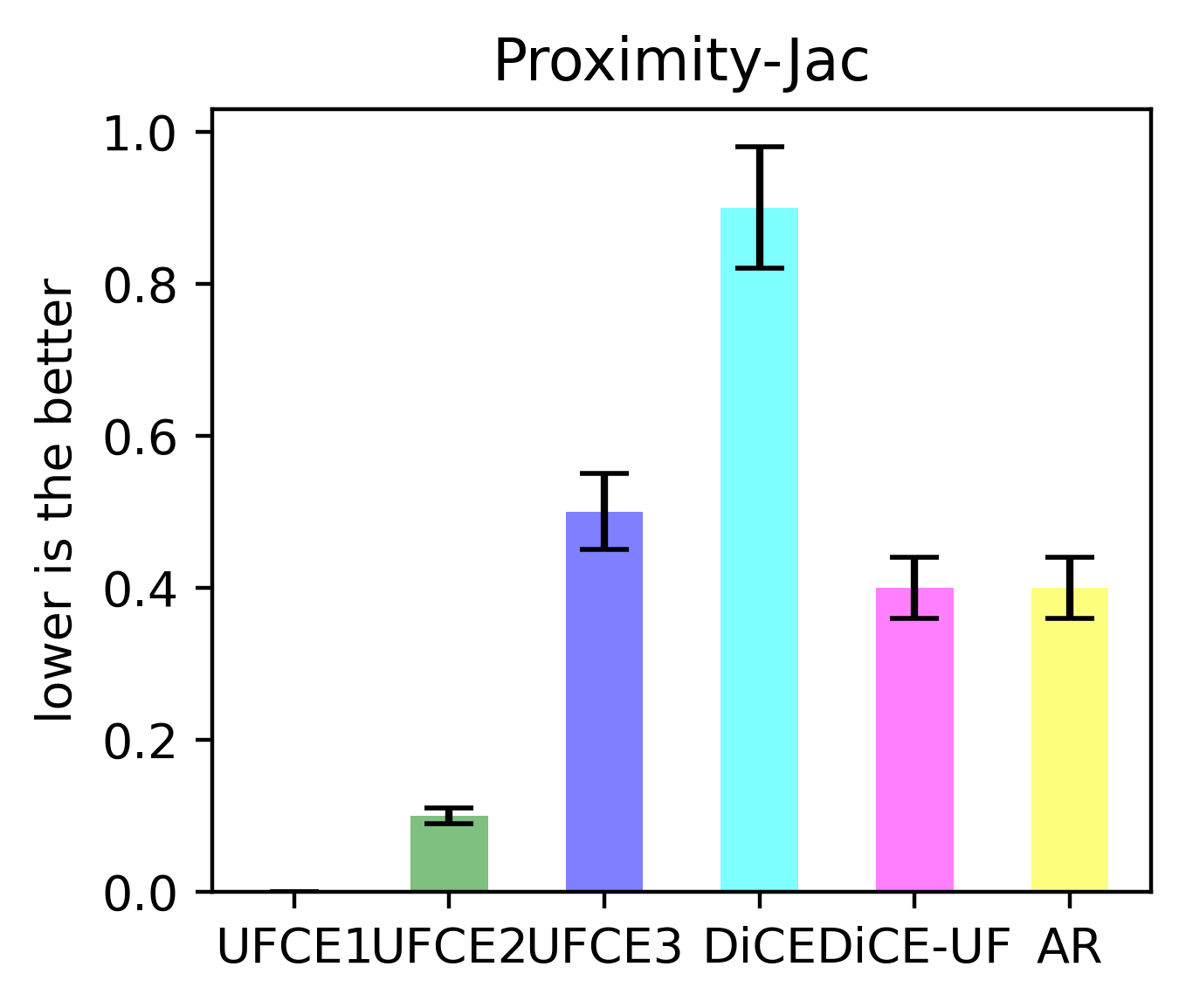}\hfill
    \includegraphics[width=.24\textwidth]{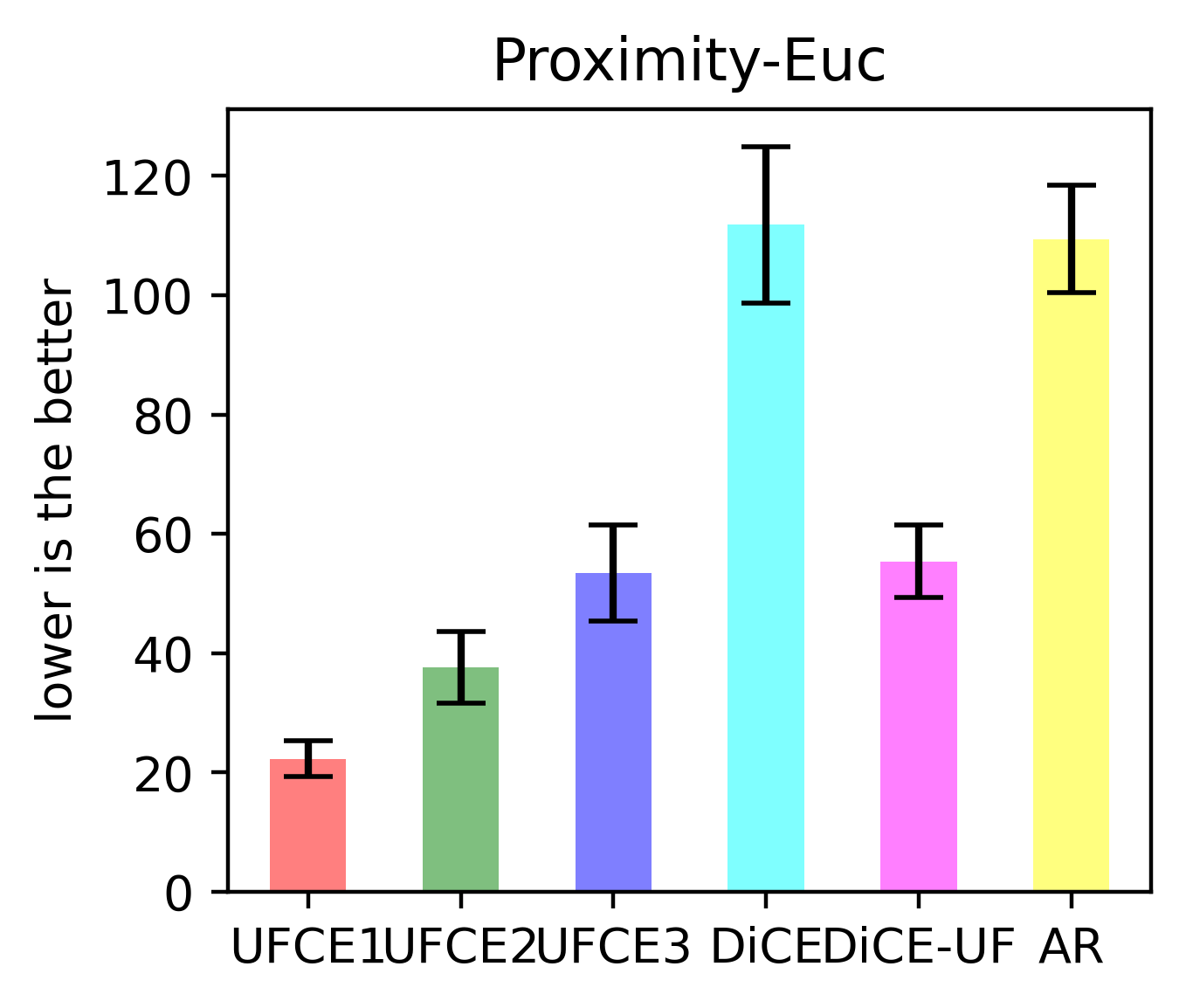}\hfill
\caption{(RQ2) Random user-preferences for CE generation: The bar plots depict the evaluation results for different evaluation metrics (with error bar of st.dev).}
\label{fig:bank-uf-mc-rq3}
\end{figure*}

The second experiment is similar to the experiment previously described in Section~\ref{sec:rq1}, the only change is in the user feedback. In this experiment, we worked with randomly taken user preferences rather than any pre-suppositions. This time we initialised the perturbation map $p$ with randomly chosen upper boundaries $p_{i^u}$ from a random choice (the lower bounds are the actual test instance values). This is not an actual \textit{Monte Carlo Simulation} but rather a random sampling of the upper bound for each feature to use in the generation of counterfactuals. For a pool of $50$ test instances, the randomly generated user feedback is utilised (in a real scenario, a user has to provide as an affordable recourse), and this process is repeated $10$ times.

The results for the feasibility metric are presented in Table~\ref{tab:stresstest}. In Table~\ref{tab:stresstest}, UFCE3 surpassed the other methods regarding the average reported results for randomly taken user constraints.
Figure~\ref{fig:bank-uf-mc-rq3} illustrates the average results for other evaluation metrics. We can observe that UFCE performed better for proximity and sparsity than other methods. We have considered the average metric scores for all the generated counterfactuals. For example, in the case of sparsity, if a method generates only $24\%$ feasible counterfactuals, then, we have computed the sparsity of the generated counterfactuals only, and taken the mean of those generated counterfactuals. In the case of UFCE1, the sparsity value is $1$, which means UFCE1 has changed only $1$ feature of the total features, whereas DiCE has changed around $6\%$ of features in average. In sparsity, UFCE1 is better than DiCE as it suggests a smaller percentage of feature changes to get CE.

The reason for better results of UFCE for feasibility (e.g., actionability) is, that it already employs the subset of features of the user-specified list of features, and perform targeted perturbations (generation of counterfactuals). Furthermore, when DiCE and AR were subjected to generate CEs on the same subset of features as for UFCE, they were only able to generate from $4\%$ to $6\%$ CEs; that being the case, DiCE and AR were allowed to exploit the rest of the features in their search for counterfactuals. It is worth noting that DiCE-UF was able to generate around $40\%$ feasible counterfactuals. To do so, DiCE-UF takes into account the given (randomly generated) user feedback similarly as UFCE does but without considering mutual information. In consequence, DiCE-UF achieved smaller percentage of feasible counterfactuals than UFCE. In  general, it was taking into account the user feedback in the generation mechanism which restricted it to not make extreme changes for finding CE. In general, UFCE (and its variations) have shown better results than DiCE and AR.

\subsection{(RQ3) What is the behaviour of UFCE on multiple datasets?}\label{sec:rq3}

We have conducted a third experiment to compare UFCE with other counterfactual methods on five datasets. This experiment follows the same set-up as the one described in Section~\ref{sec:rq1}. Nevertheless, the user constraints are now fixed to a threshold equal to $50\%$ of the median absolute deviations (MAD) of features in the actual data distribution for each test fold. Each dataset was split into 5-test folds, and the mean results of CE generation for all folds are reported. 

\begin{figure*}[hbt!]
\begin{subfigure}[t]{0.24\textwidth}
    \makebox[0pt][r]{\makebox[15pt]{\raisebox{15pt}{\rotatebox[origin=c]{90}{Graduate}}}}%
    \includegraphics[width=\textwidth]
    {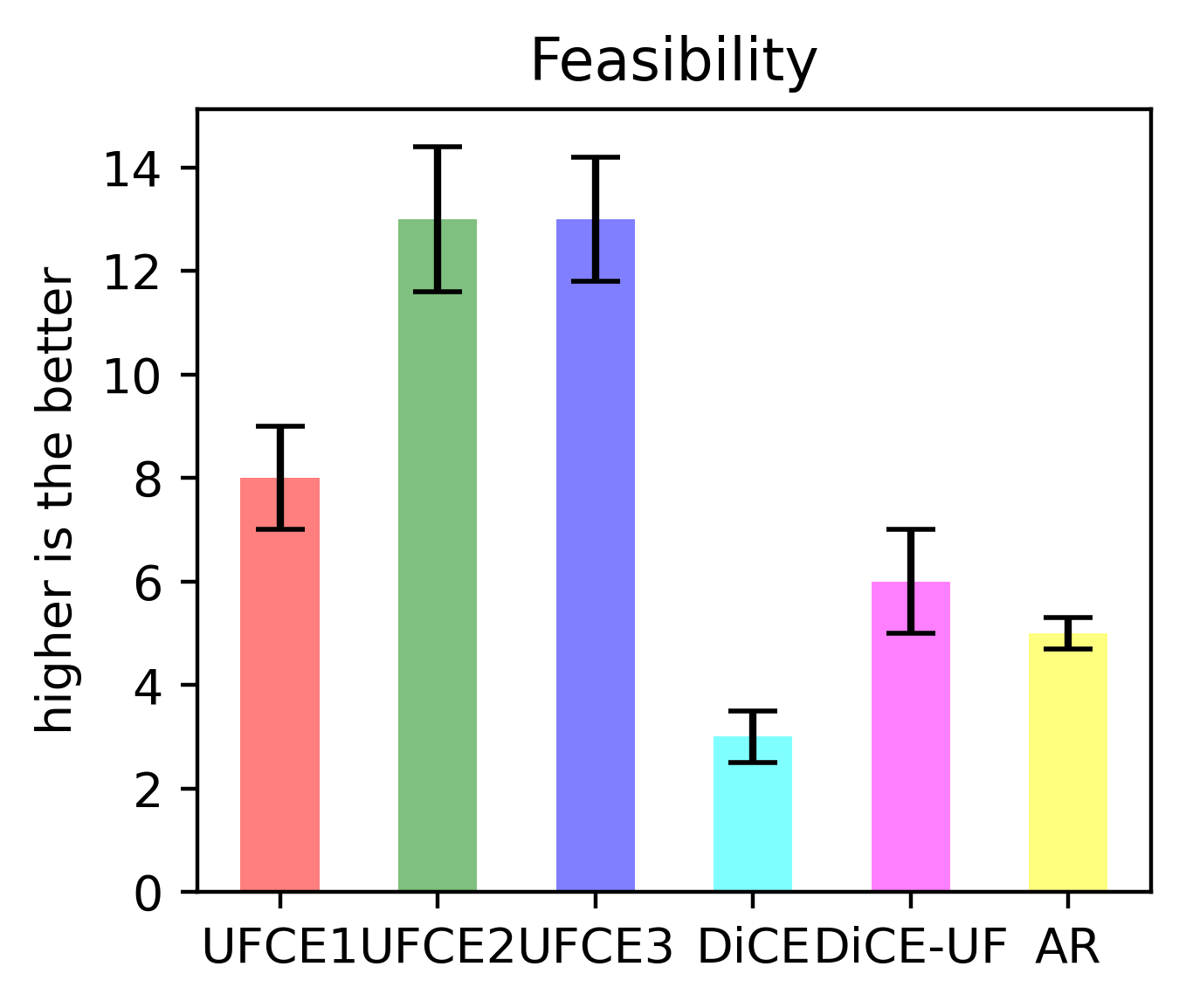}\hfill
 \makebox[0pt][r]{\makebox[15pt]{\raisebox{15pt}{\rotatebox[origin=c]{90}{Bank}}}}%
    \includegraphics[width=\textwidth]
    {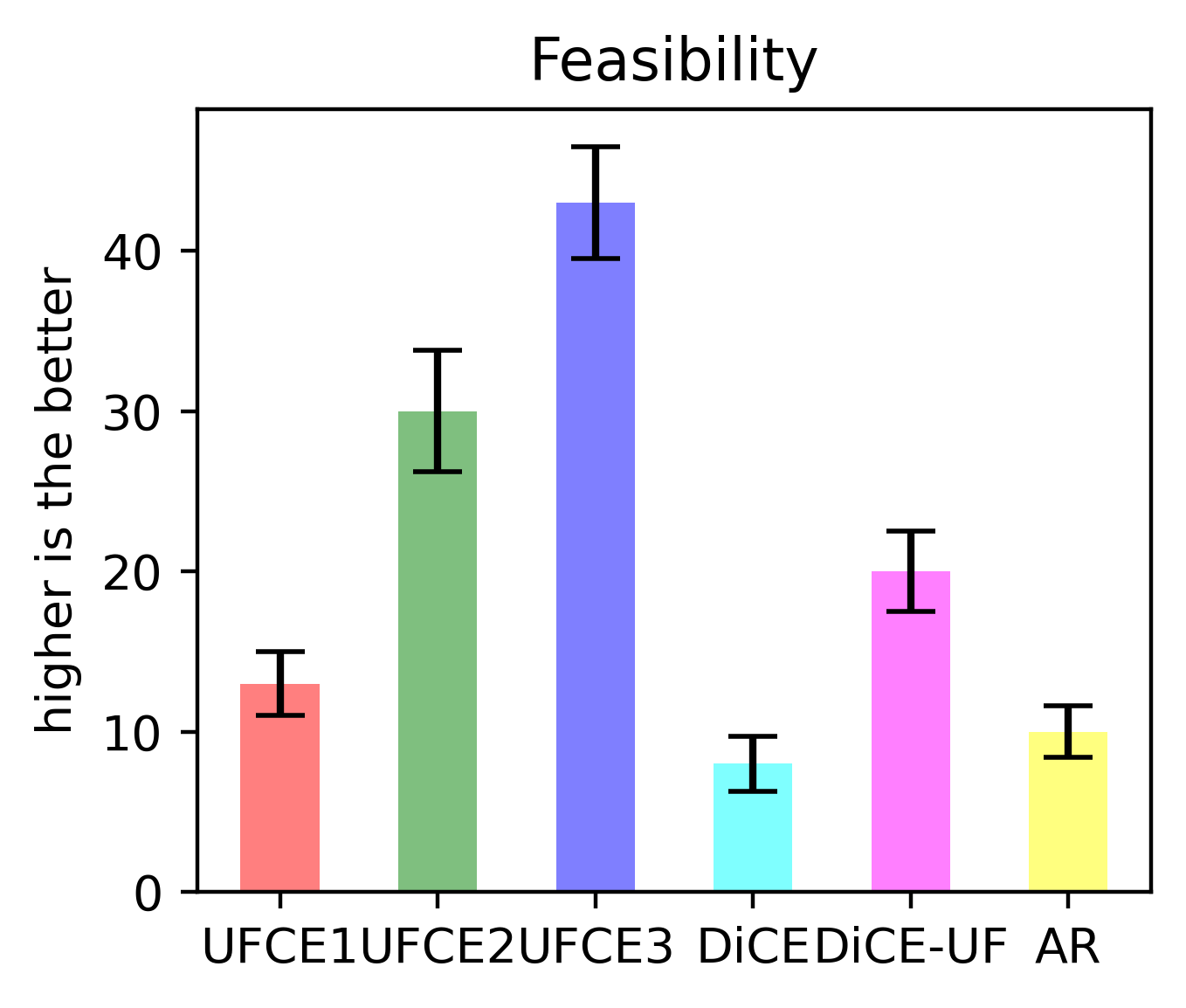}\hfill
     \makebox[0pt][r]{\makebox[15pt]{\raisebox{15pt}{\rotatebox[origin=c]{90}{Movie}}}}%
    \includegraphics[width=\textwidth]
    {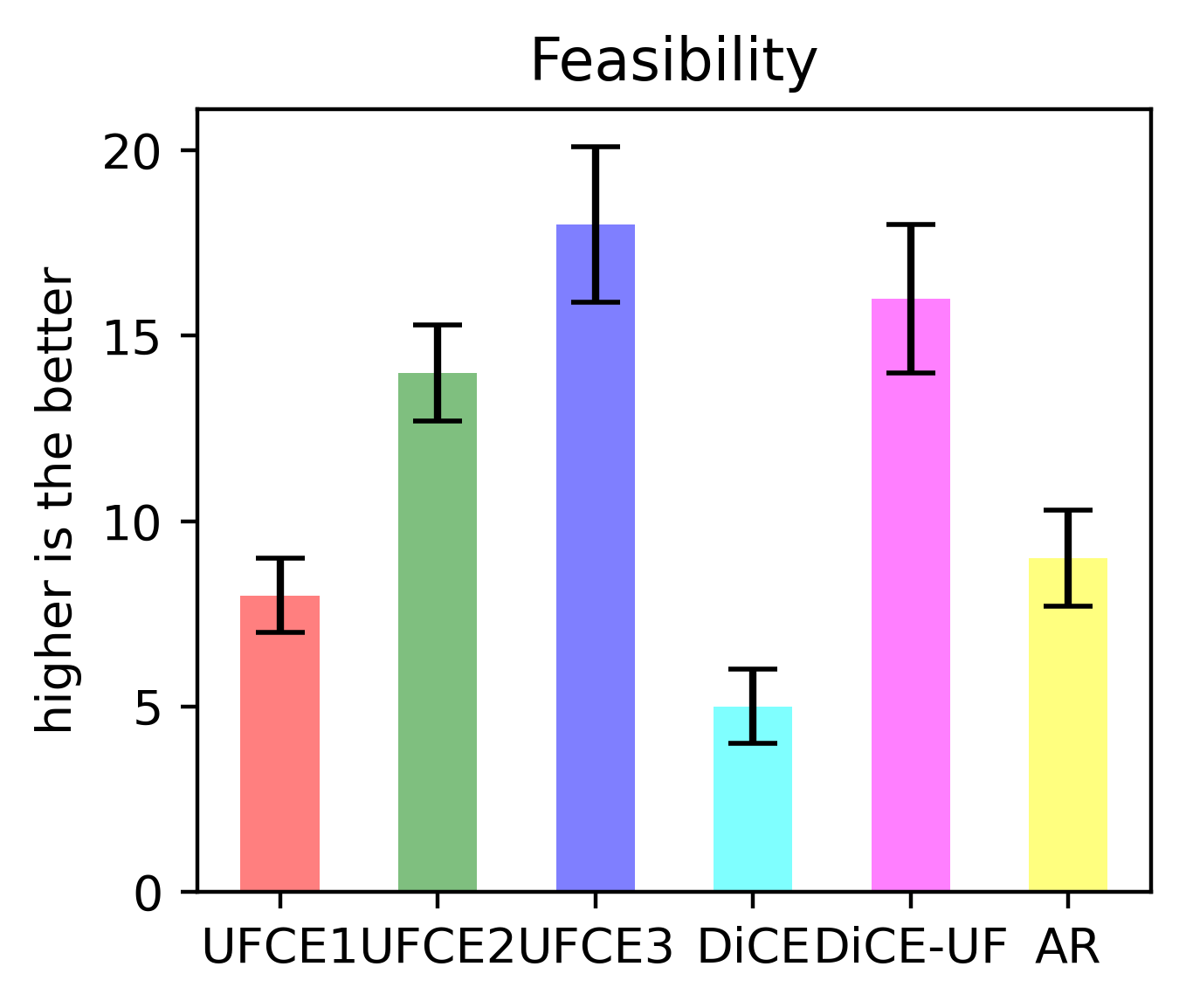}\hfill
     \makebox[0pt][r]{\makebox[15pt]{\raisebox{15pt}{\rotatebox[origin=c]{90}{Wine}}}}%
    \includegraphics[width=\textwidth]
    {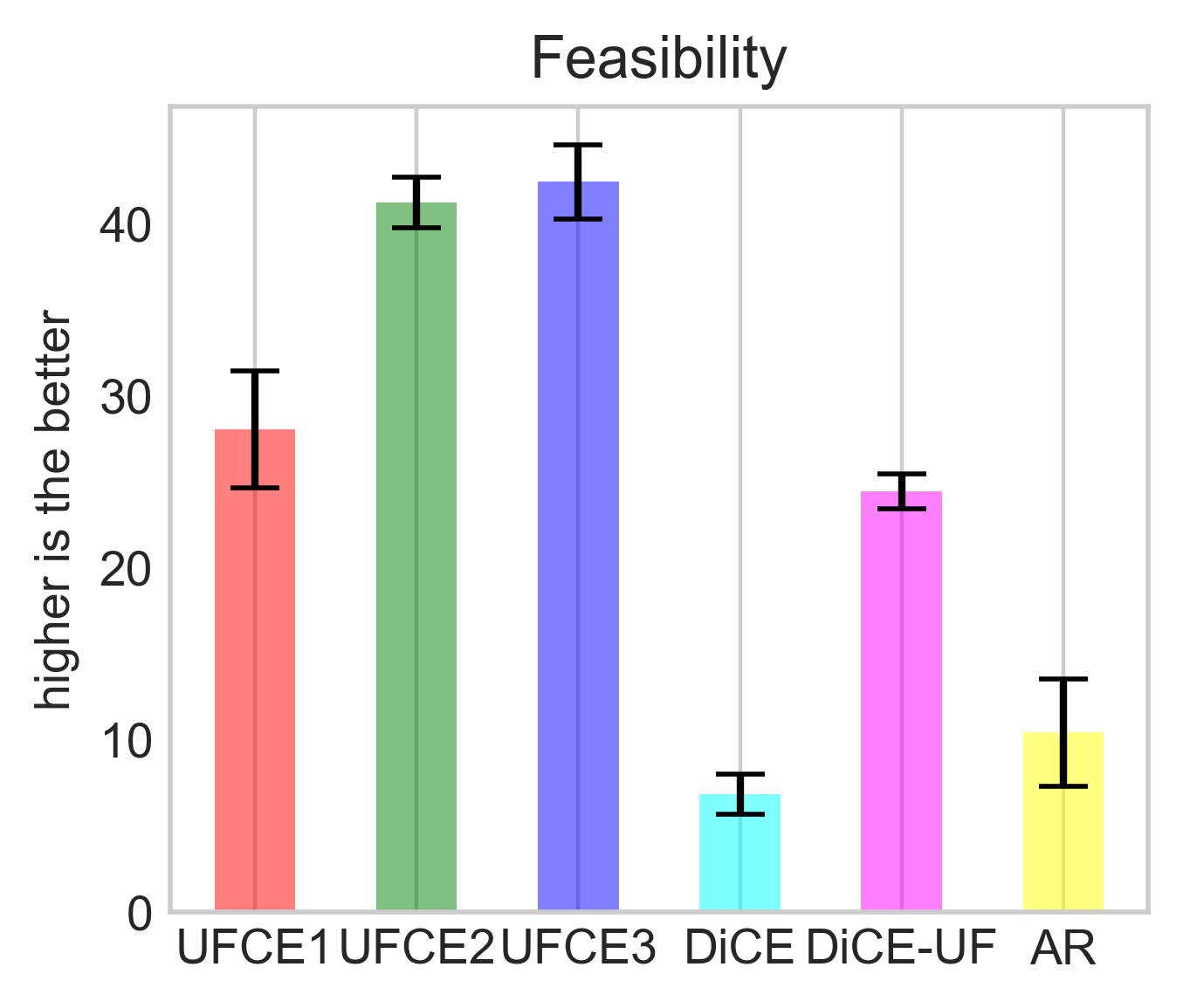}\hfill
     \makebox[0pt][r]{\makebox[15pt]{\raisebox{15pt}{\rotatebox[origin=c]{90}{Bupa}}}}%
    \includegraphics[width=\textwidth]
    {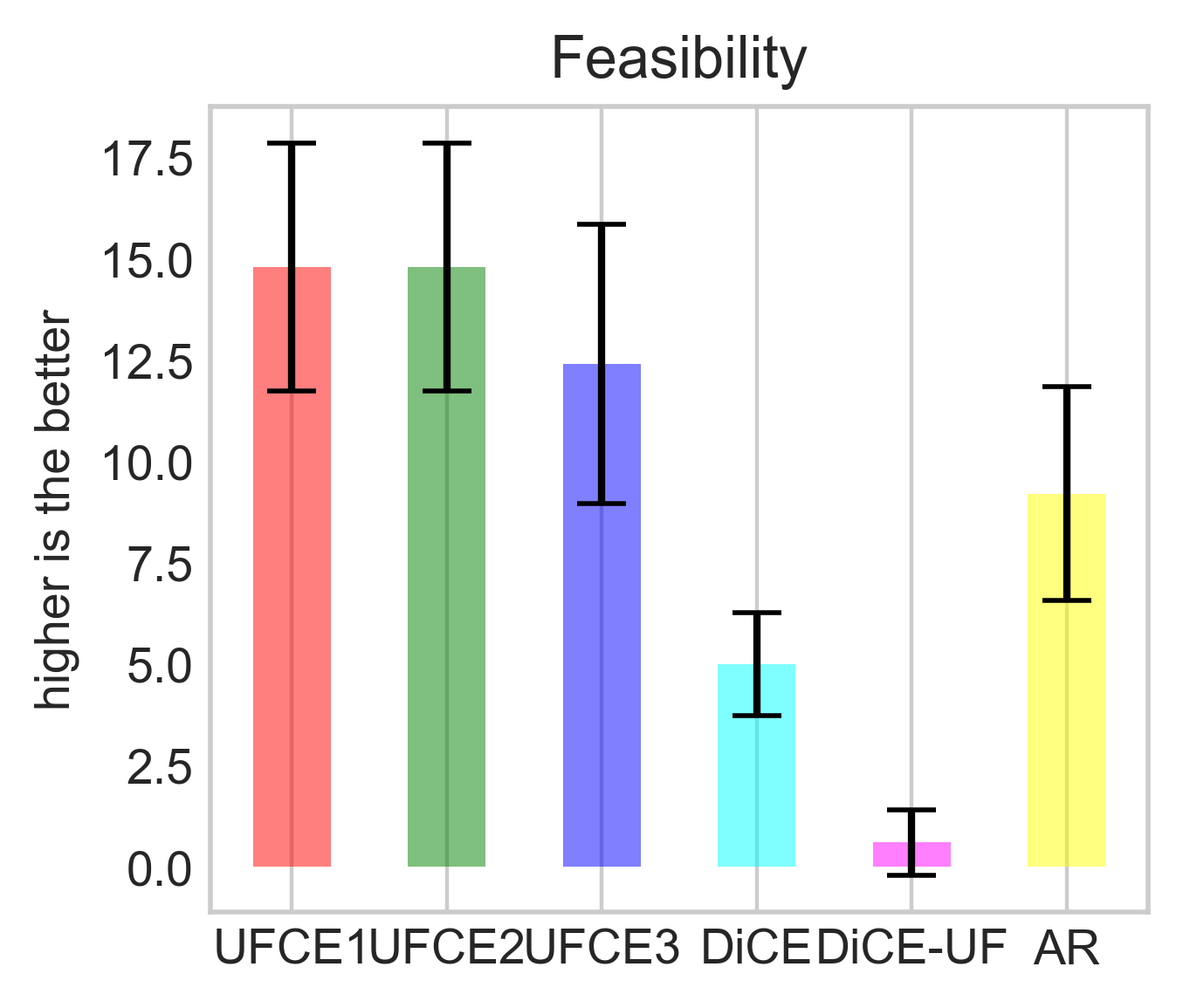}\hfill
\end{subfigure}
\begin{subfigure}[t]{0.24\textwidth}
    \makebox[0pt][r]{\makebox[15pt]{\raisebox{15pt}{\rotatebox[origin=c]{90}{}}}}%
    \includegraphics[width=\textwidth]
    {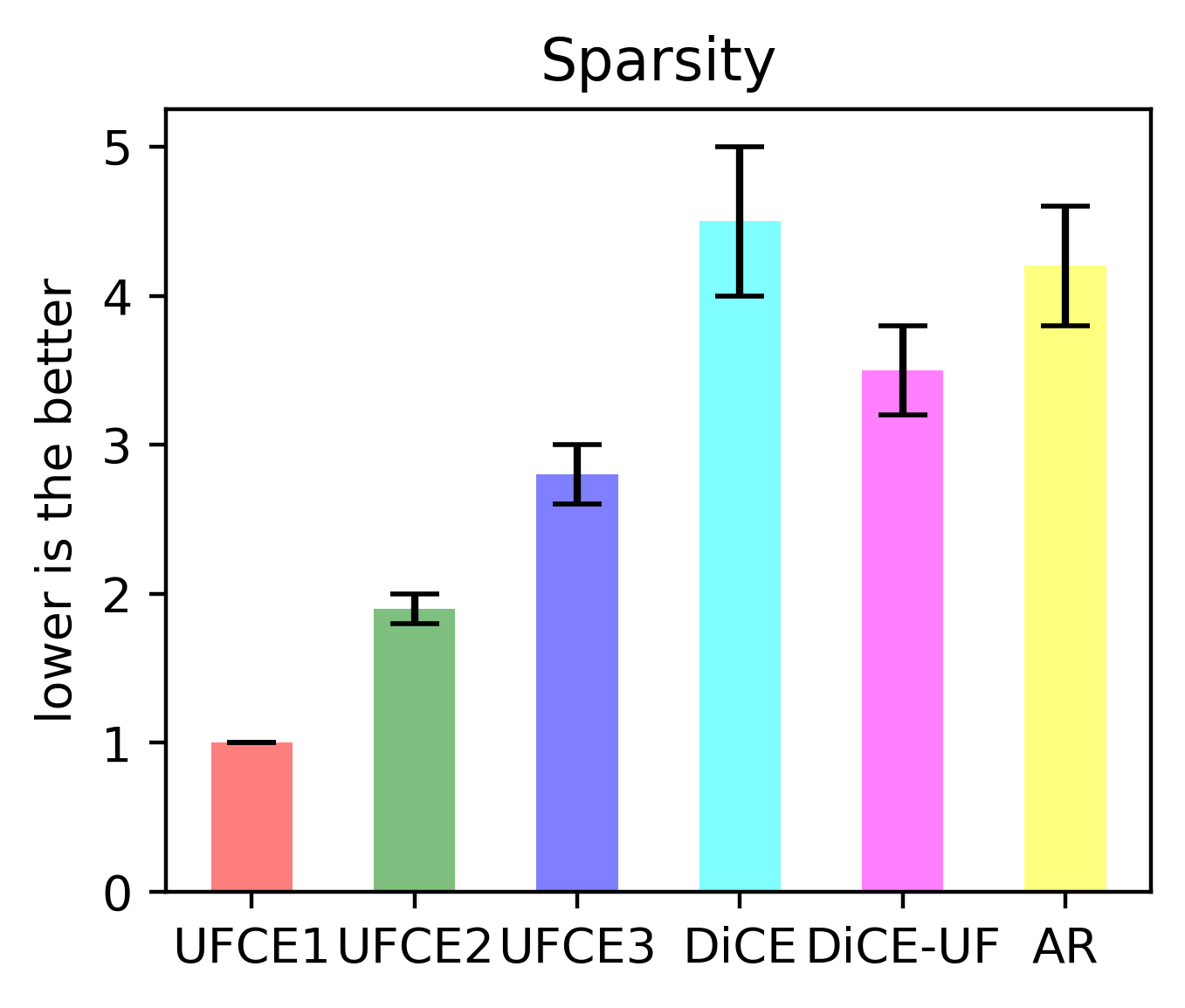}\hfill
 \makebox[0pt][r]{\makebox[15pt]{\raisebox{15pt}{\rotatebox[origin=c]{90}{}}}}%
    \includegraphics[width=\textwidth]
    {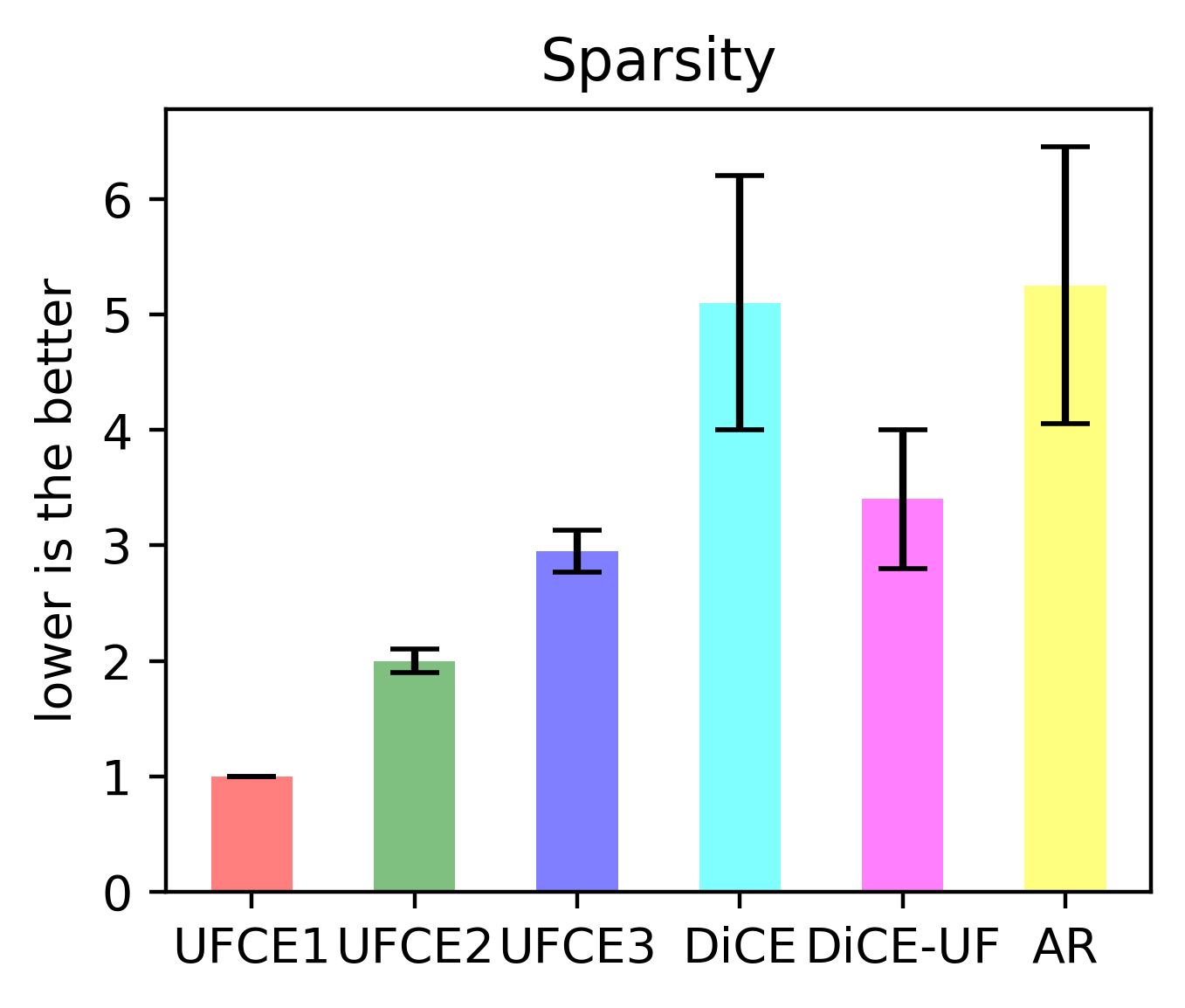}\hfill
     \makebox[0pt][r]{\makebox[15pt]{\raisebox{15pt}{\rotatebox[origin=c]{90}{}}}}%
    \includegraphics[width=\textwidth]
    {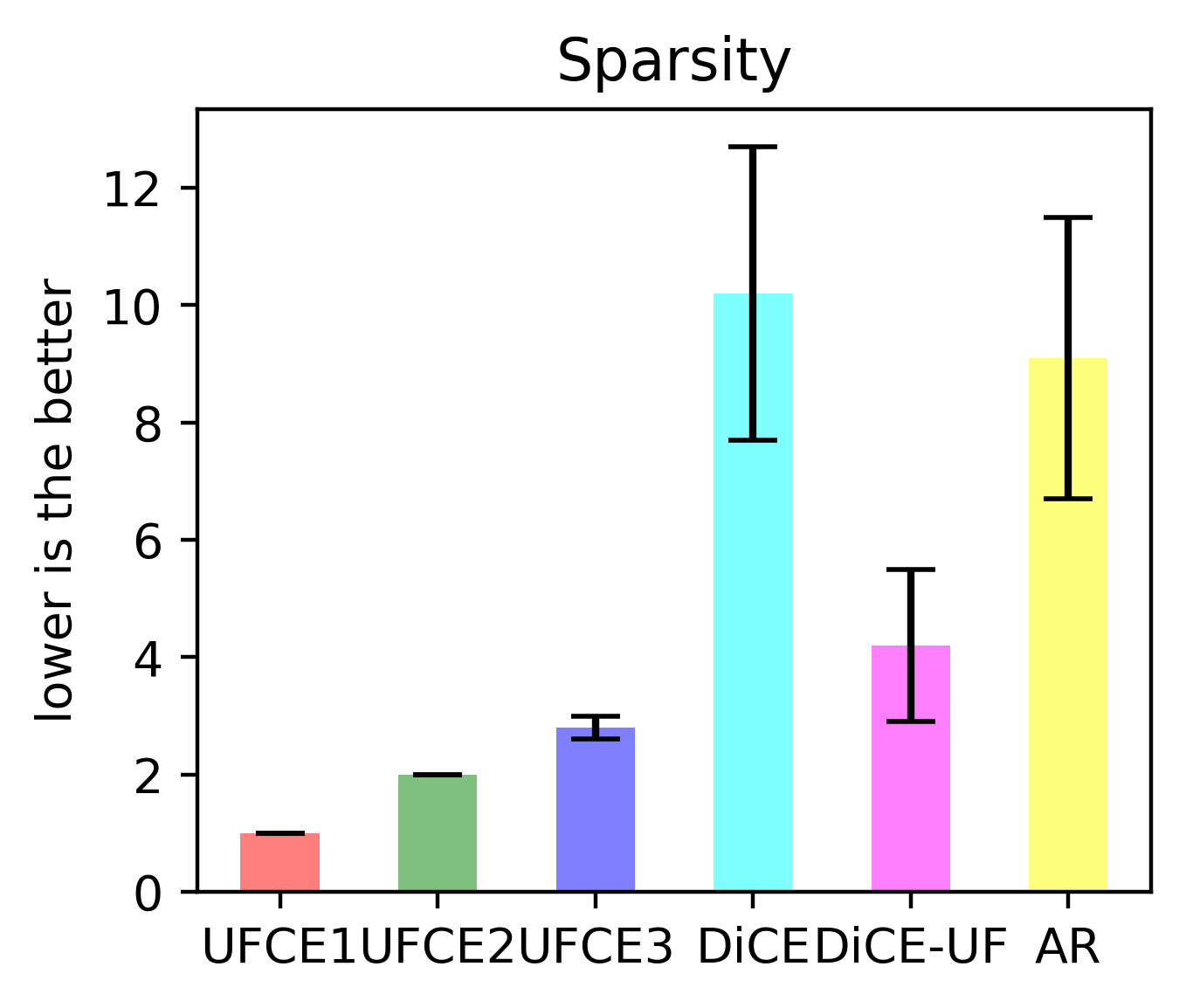}\hfill
     \makebox[0pt][r]{\makebox[15pt]{\raisebox{15pt}{\rotatebox[origin=c]{90}{}}}}%
    \includegraphics[width=\textwidth]
    {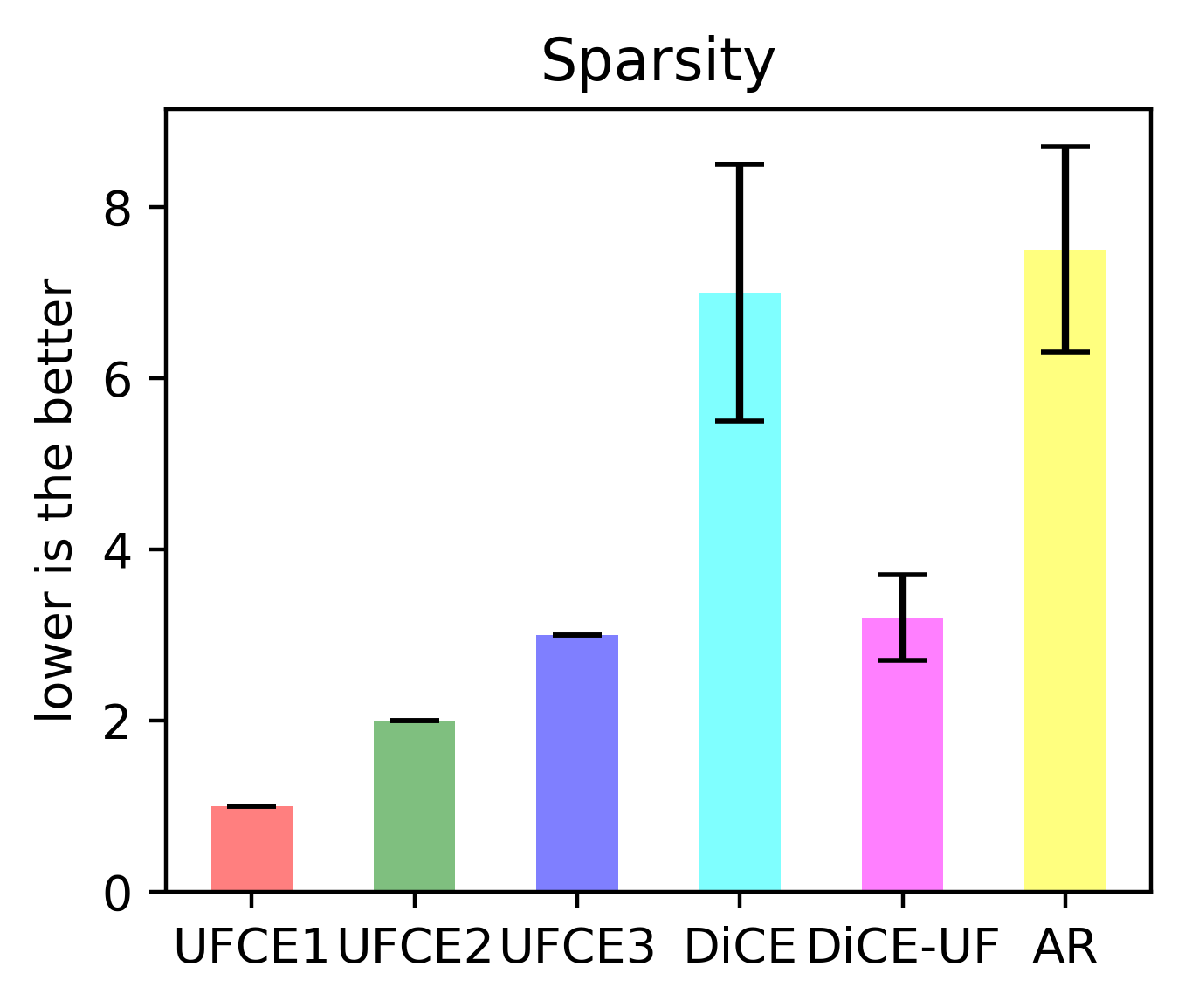}\hfill
     \makebox[0pt][r]{\makebox[15pt]{\raisebox{15pt}{\rotatebox[origin=c]{90}{}}}}%
    \includegraphics[width=\textwidth]
    {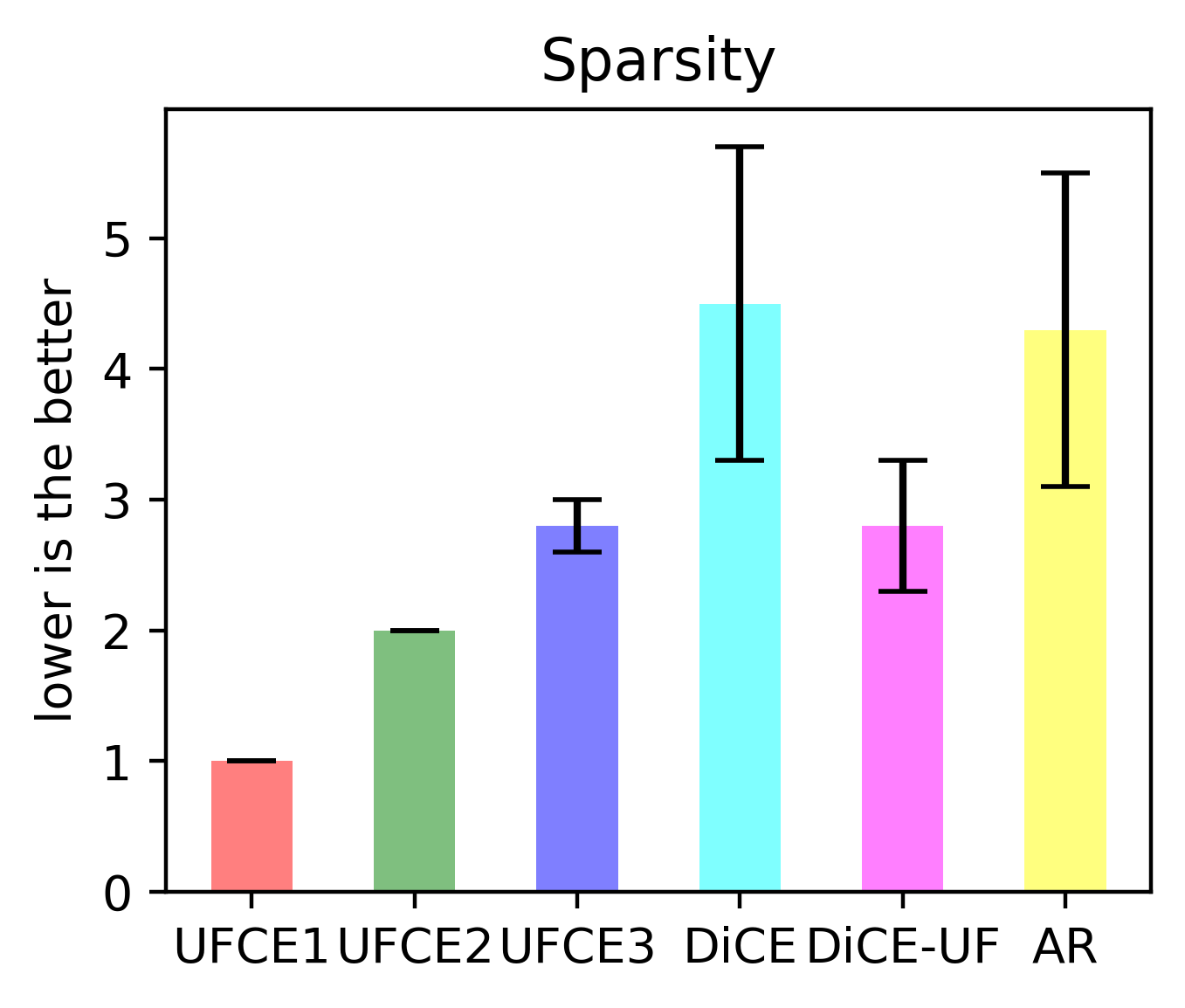}\hfill
\end{subfigure}
\begin{subfigure}[t]{0.24\textwidth}
    \makebox[0pt][r]{\makebox[15pt]{\raisebox{15pt}{\rotatebox[origin=c]{90}{}}}}%
    \includegraphics[width=\textwidth]
    {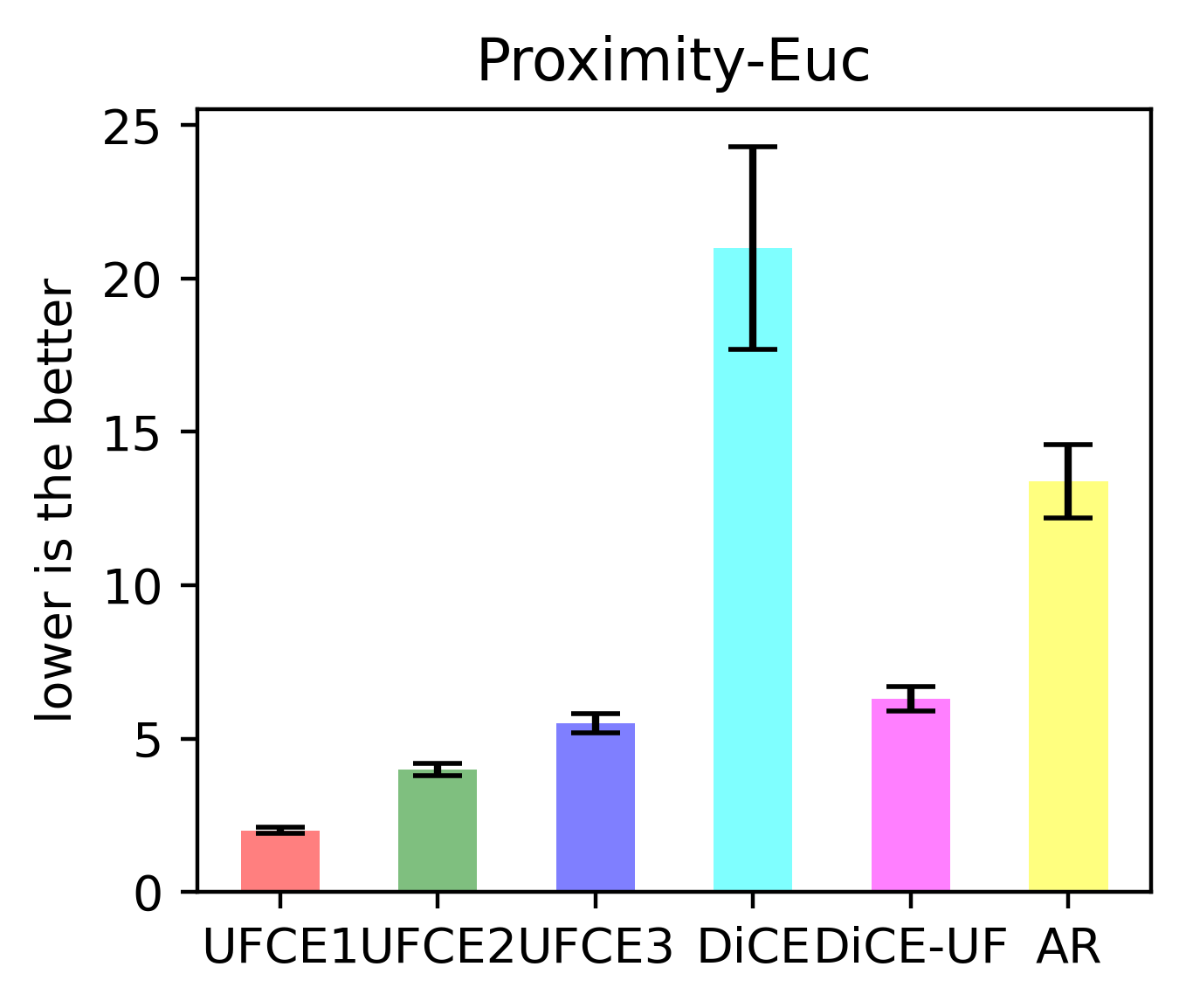}\hfill
 \makebox[0pt][r]{\makebox[15pt]{\raisebox{15pt}{\rotatebox[origin=c]{90}{}}}}%
    \includegraphics[width=\textwidth]
    {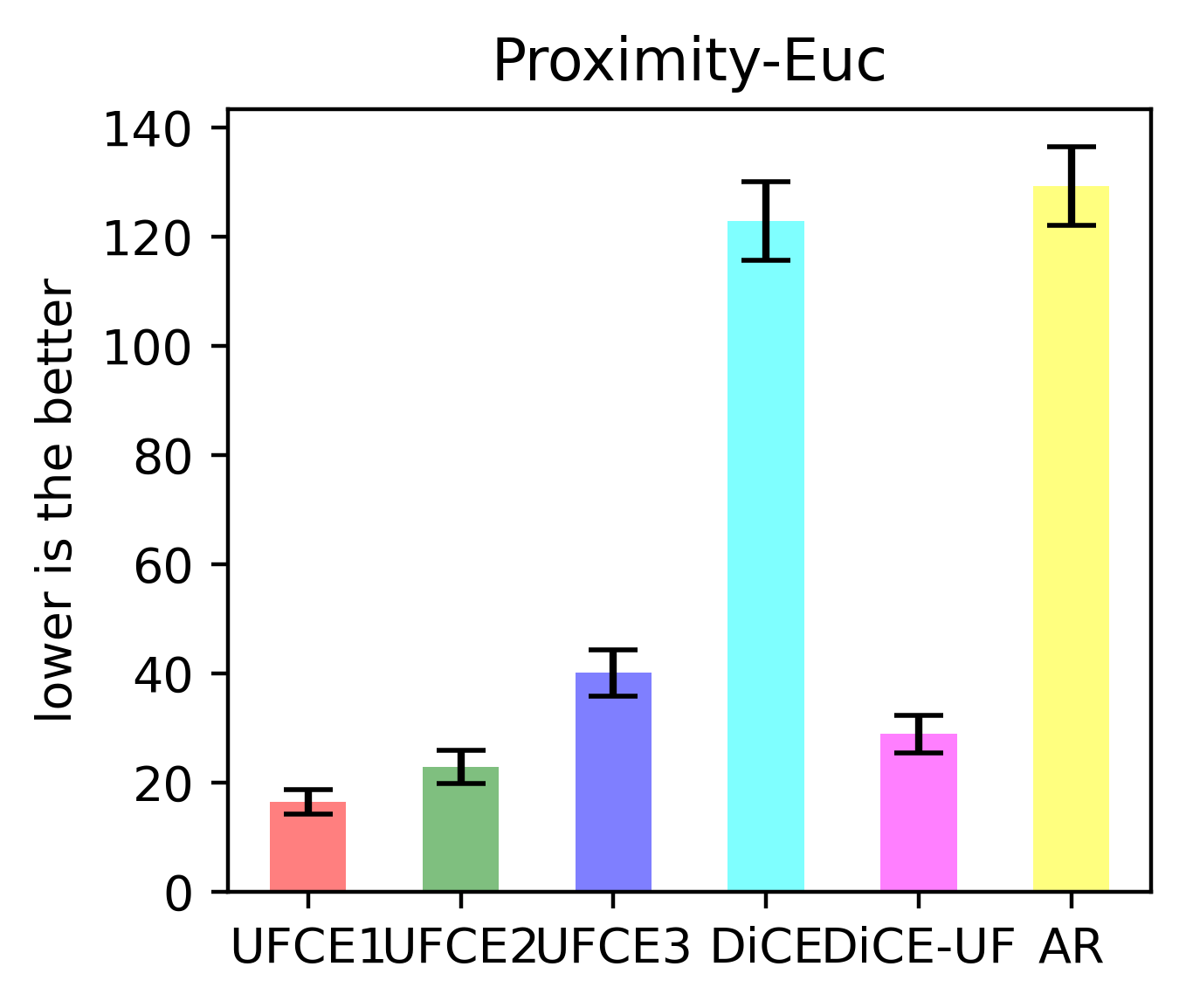}\hfill
     \makebox[0pt][r]{\makebox[15pt]{\raisebox{15pt}{\rotatebox[origin=c]{90}{}}}}%
    \includegraphics[width=\textwidth]
    {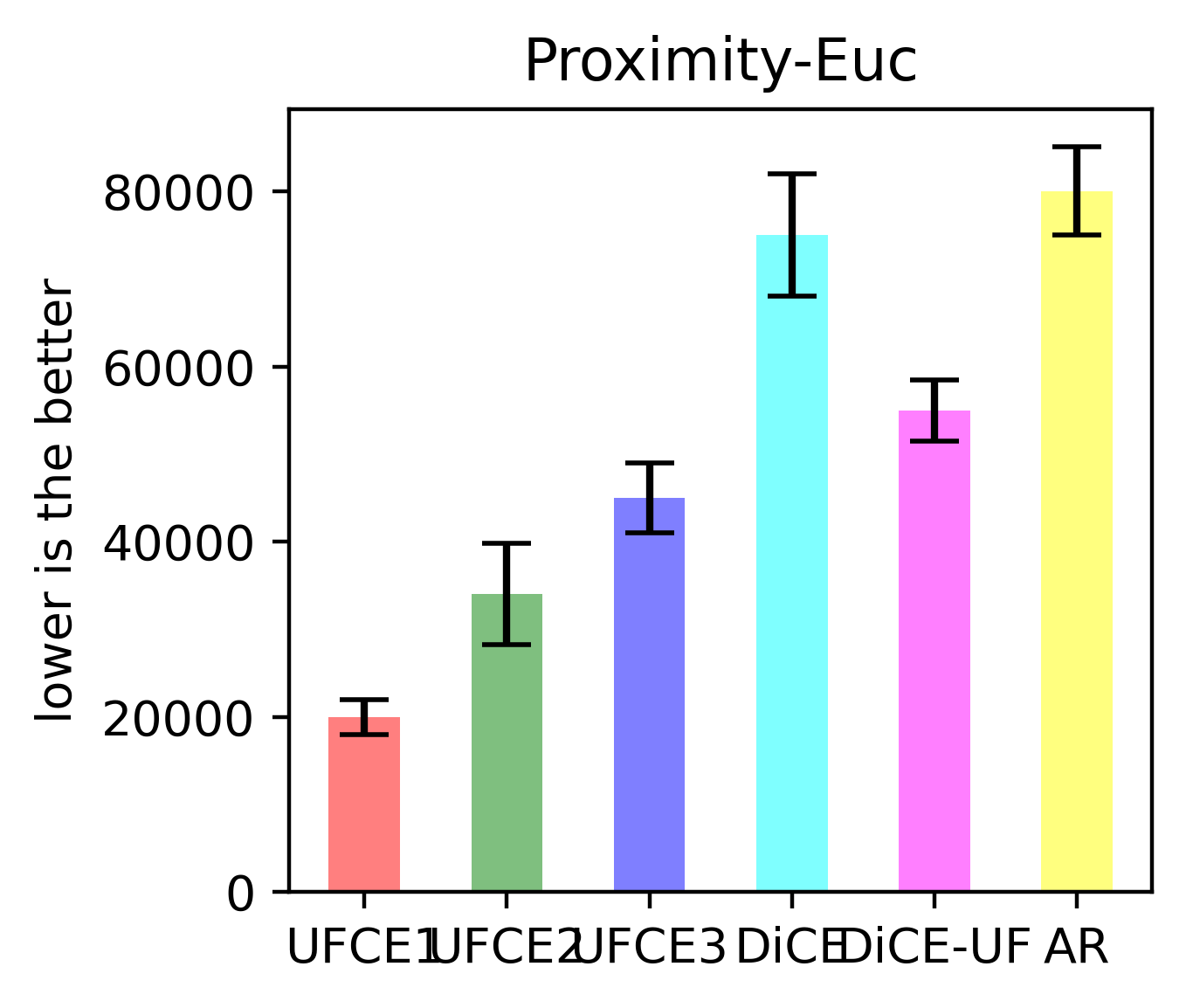}\hfill
     \makebox[0pt][r]{\makebox[15pt]{\raisebox{15pt}{\rotatebox[origin=c]{90}{}}}}%
    \includegraphics[width=\textwidth]
    {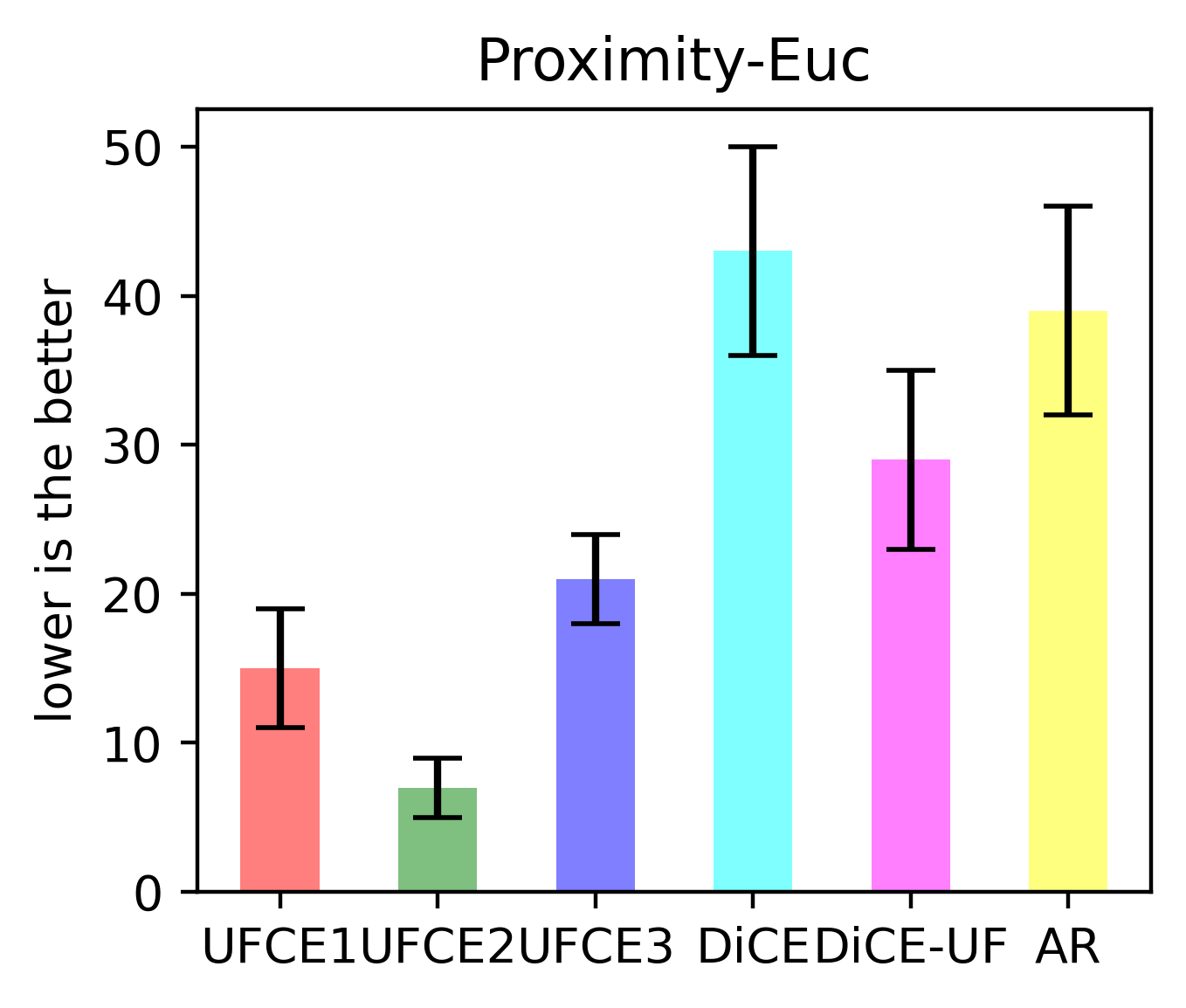}\hfill
     \makebox[0pt][r]{\makebox[15pt]{\raisebox{15pt}{\rotatebox[origin=c]{90}{}}}}%
    \includegraphics[width=\textwidth]
    {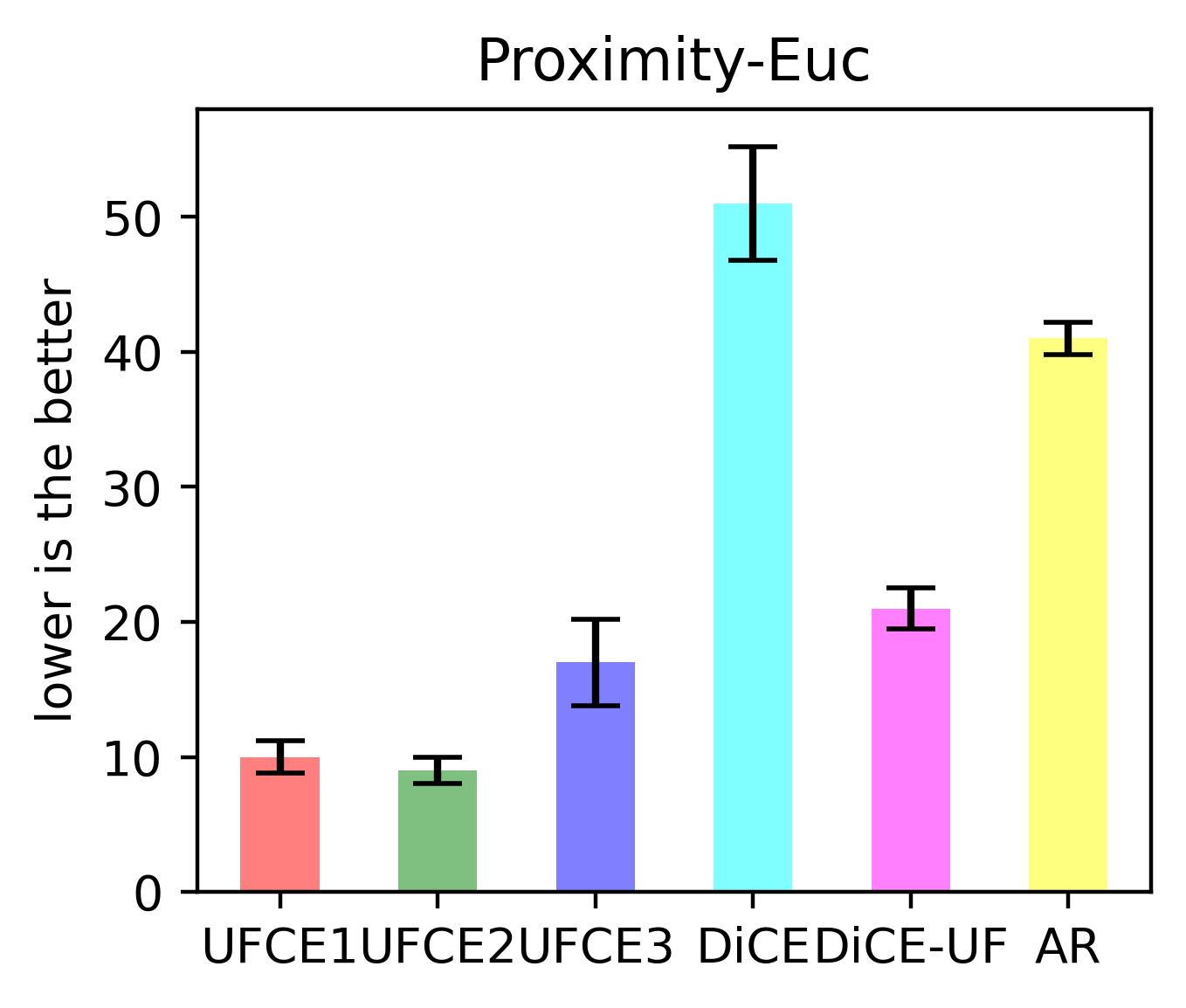}\hfill
\end{subfigure}
\begin{subfigure}[t]{0.24\textwidth}
    \makebox[0pt][r]{\makebox[15pt]{\raisebox{15pt}{\rotatebox[origin=c]{90}{}}}}%
    \includegraphics[width=\textwidth]
    {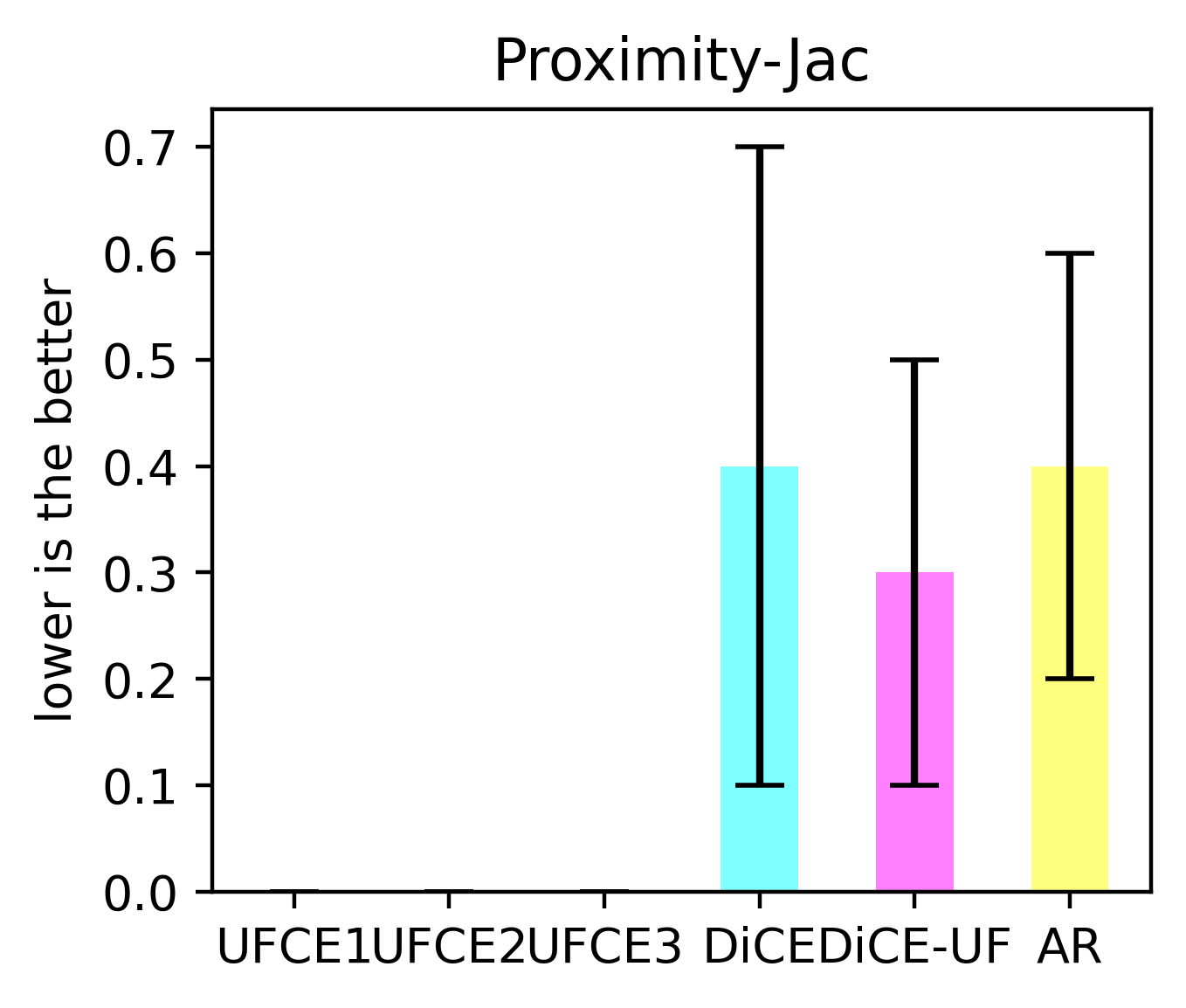}\hfill
 \makebox[0pt][r]{\makebox[15pt]{\raisebox{15pt}{\rotatebox[origin=c]{90}{}}}}%
    \includegraphics[width=\textwidth]
    {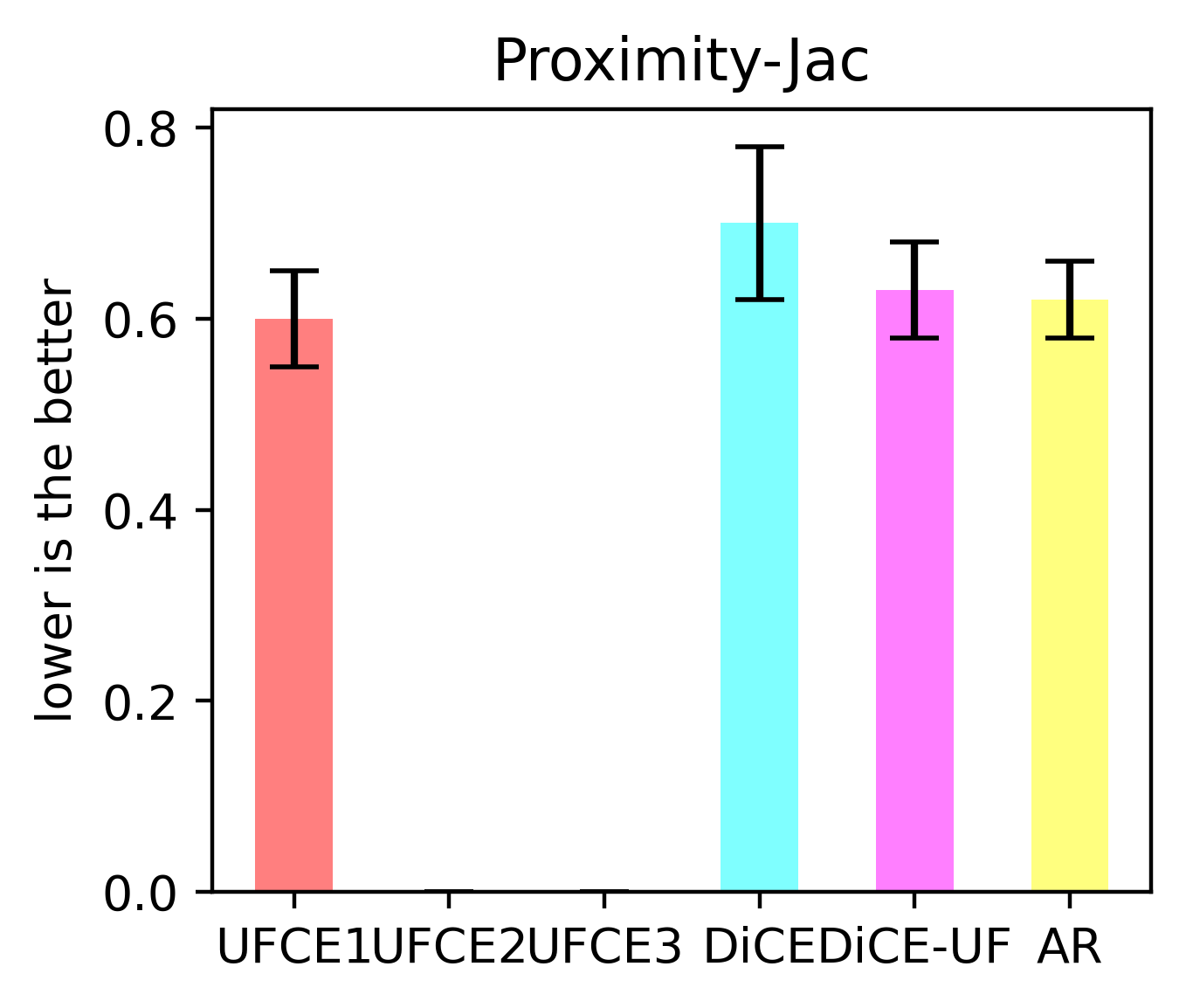}\hfill
     \makebox[0pt][r]{\makebox[15pt]{\raisebox{15pt}{\rotatebox[origin=c]{90}{}}}}%
    \includegraphics[width=\textwidth]
    {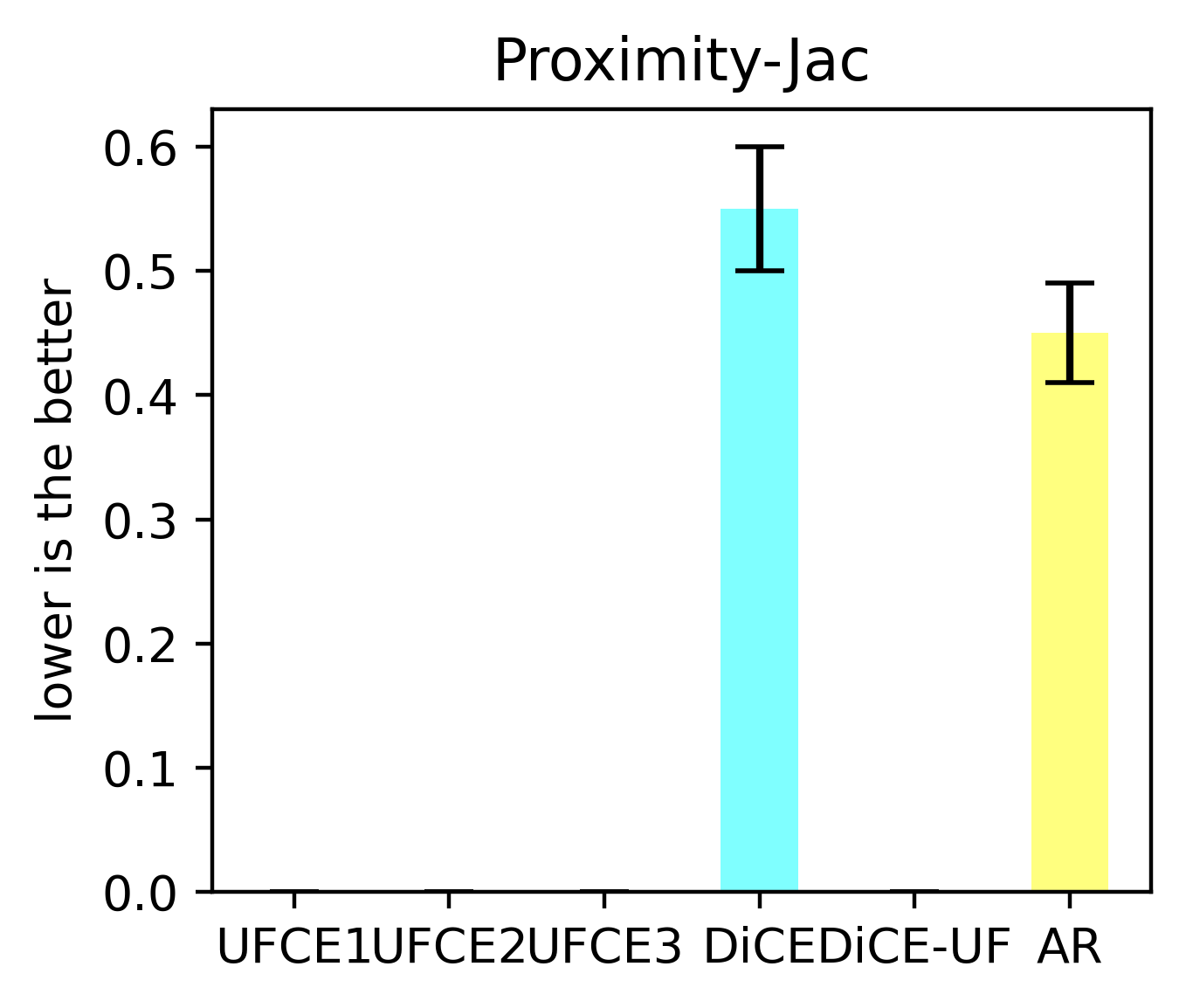}\hfill
\end{subfigure}
\caption{(RQ3) Comparative results for CE generation on multiple datasets: The bar plots depict the evaluation results for different evaluation metrics (with error bar of st.dev).}
\label{fig:rq3}
\end{figure*}

Table~\ref{tab:datasets} (introduced in Section~\ref{sec:rq1}) contains the details about the different datasets, their features, and the ML model's 5-fold cross-validation (CV) mean accuracy. 
The comparative results (mean of different folds of test set) on five datasets are presented for proximity-Jac, proximity-Euc, sparsity, actionability, plausibility, and feasibility in Table~\ref{tab:comparative-rq2}. Fig.~\ref{fig:rq3} provides the readers with complementary bar plots to facilitate interpretation of numbers reported in Table~\ref{tab:comparative-rq2}.

\begin{table*}[t!]
\caption{(RQ3) Comparative results on multiple datasets for different evaluation metrics. The evaluation metrics are provided with up-arrow $\uparrow$ to show that higher is better and down-arrow $\downarrow$ for lower is better. The $na$ denotes not applicable (in datasets where categorical features are not present).}
\label{tab:comparative-rq2}
\resizebox{\textwidth}{!}{
\begin{tabular}{p{0.09\textwidth}p{0.09\textwidth}p{0.10\textwidth}p{0.10\textwidth}p{0.09\textwidth}p{0.12\textwidth}p{0.11\textwidth}p{0.11\textwidth}}
\toprule
dataset & CE Method & prox-Jac $\downarrow$ & prox-Euc $\downarrow$ & sparsity $\downarrow$ & actionability $\uparrow$ & plausibility $\uparrow$ & feasibility $\uparrow$ \\ \midrule
\multirow{5}{*}{Graduate} 
& DiCE & 0.40 & 22.12 & 4.40  & 3.00  & 9.00 & 3.00 \\
& DiCE-UF & 0.30 & 6.32 & 3.40 & 6.00 & 10.00 & 6.00 \\
& AR       & 0.40  & 13.34  & 4.15 & 5.00 & 9.00 & 5.00 \\
& UFCE1  & \textbf{0.00} & \textbf{2.34} & \textbf{1.00} & 8.00 & 8.00 & 8.00 \\ 
& UFCE2 & 0.00 & 4.85 & 2.00 & \textbf{13.00} & \textbf{13.00} & \textbf{13.00} \\ \midrule
\multirow{5}{*}{BankLoan} 
& DiCE       & 0.70  & 121.45 & 5.20 & 8.00 & 39.00 & 8.00 \\
& DiCE-UF & 0.65 & 26.12 & 3.20 & 34.00 & 20.00 & 20.00 \\
& AR          & 0.62 & 129.56 & 5.30 & 9.00 & \textbf{45.00} & 9.00 \\
& UFCE1  & 0.60 & \textbf{10.00} & \textbf{1.00} & 14.00 & 14.00 & 14.00  \\
& UFCE2  & \textbf{0.00} & 23.10 & 2.00 & 30.00 & 30.00 & 30.00  \\ \midrule
\multirow{6}{*}{Wine} 
& DiCE        & na  & 43.10 & 7.00 & 10.00  & 33.00 & 10.00 \\
& DiCE-UF  & na & 28.15 & 3.00 & \textbf{50.00} & 25.00 & 25.00 \\
& AR           & na  & 38.25 & 7.20 & 11.00  & 42.00  & 11.00 \\
& UFCE1 & na  & 14.90 & \textbf{1.00} & 43.00  & 28.00 & 28.00 \\ 
& UFCE2 & na & \textbf{8.45} & 2.00  & \textbf{50.00} & 41.00 & 41.00 \\ 
& UFCE3 & na  & 21.95  & 3.00  & \textbf{50.00}  & \textbf{42.00} & \textbf{42.00} \\ \midrule
\multirow{6}{*}{Bupa} 
& DiCE &      na & 51.45 & 4.50  & 1.00  & 1.00  & 1.00 \\
& DiCE-UF & na & 20.34 & 2.90  & 5.00  & 12.00 & 5.00 \\
& AR         & na & 40.65 & 4.20 & 9.00 & \textbf{15.00} & 9.00\\
& UFCE1 & na & 10.00 & \textbf{1.00} & \textbf{17.00}  & \textbf{15.00}  & \textbf{15.00}   \\ 
& UFCE2 & na & \textbf{9.00}  & 2.00 & 15.00 & \textbf{15.00} & \textbf{15.00}   \\ 
& UFCE3 & na & 17.10 & 2.90 & 13.00 & 13.00  & 13.00   \\ \midrule
\multirow{6}{*}{Movie} 
& DiCE  &  0.55 & 78.00 & 10.00 & 5.00 & \textbf{19.00}  & 5.00  \\
& DiCE-UF & 0.00 & 56.00 & 4.00 & \textbf{20.00}  & 17.00  & 17.00  \\
& AR         & 0.45  & 80.00 & 9.00 & 8.00 & \textbf{19.00} & 8.00 \\
& UFCE1 & \textbf{0.00} & \textbf{20.00} & \textbf{1.00} & \textbf{20.00}  & 8.00 & 8.00 \\ 
& UFCE2 & 0.00 & 32.00 & 2.00 & 14.00 & 14.00  & 14.00 \\ 
& UFCE3 & 0.00 & 43.00 & 3.00 & 18.00 & 18.00  & \textbf{18.00}  \\ \midrule 
\end{tabular}%
}
\end{table*}

We can observe that UFCE performed better in most of the evaluation metrics on multiple datasets. The better result from any of the methods on any dataset for each specific evaluation metric is highlighted in bold in Table~\ref{tab:comparative-rq2}. Regarding proximity-Jac, there are three datasets which have categorical features, UFCE1 utilised $0.6 (60\%)$ of the categorical features for Bank loan dataset. For Graduate and Movie datasets, no UFCE variation utilised categorical features. This is positive in a sense that the user has not to change the category of the features which in some cases is not viable like to change the \textit{gender} feature in some real world dataset. Regrading the proximity-Euc, UFCE1 performed better than others on Graduate, Bank Loan and Movie datasets, while UFCE2 performed better than others on Wine and Bupa datasets. Regarding sparsity, UFCE1 performed better than all other methods by suggesting only one feature change. Regarding actionability, UFCE1 and DiCE-UF shared the best performnce on Movie dataset; UFCE1 performed better than others on Bupa dataset; UFCE3 performed better than others on Graduate, Bank Loan, and Wine datasets. Regarding plausibility, UFCE2 and UFCE3 performed better than others on Graduate dataset; AR performed better than others on Bank Loan dataset; AR and UFCE3 shared the best performance on Wine dataset; AR, UFCE1, and UFCE2 shared the better performance than others on Bupa dataset; AR and DiCE shared the better performance than others on Movie dataset. Regarding feasibility, UFCE2 and UFCE3 shared the best performance on Graduate dataset; UFCE3 performed better than others on Bank Loan, Wine, and Movie datasets; UFCE1 and UFCE2 shared the better performance than others on Bupa dataset.

Finally, we can draw some conclusions about how well the various UFCE variations have done regarding proximity and sparsity. The generated counterfactuals are meaningful and easy to understand because they are situated relatively near to the described test cases, and there are a few modifications. UFCE produces coherent counterfactuals while adhering to user-defined actionability restrictions. The counterfactuals produced by UFCE are based on the distribution of data from the same class of ground truth, which has a high plausibility score. This ensures that the created counterfactuals are plausible and consequently feasible.
Accordingly, UFCE consistently exhibited better results across all the five datasets under study. These positive outcomes suggested robustness and effectiveness in various scenarios. Acknowledging the nuanced nature of performance evaluation, these findings provide promising indications of the efficacy of UFCE across similar tabular datasets.
\section{Conclusion and Future Work}
\label{conclusion}
Even though the rules governing interpretable algorithms are still in their early stages, regulations demand explanations ensuring actionable information fulfilling human needs. The customers of explainable systems have been empowered by laws to get actionable information. In specific domains, for example, in credit scoring, to make the customers aware of adverse actions, an Act is designed for Equal Credit Opportunity in United States \cite{equalAct}. Our approach strives to provide actionable information by involving the user to gain the utmost trust in the generated CEs. The reveal of actionable information could benefit the domain experts in debugging and diagnosis. In contrast, reverse engineering could be applied using actionable information to learn the model's behaviour (model internals, which the model owners never want to reveal). We tried to balance this trade-off confining the user to customise the information to some extent while respecting their rights to explanations.

This study introduces a novel methodology (UFCE) for generating user feedback-based CEs, which addresses the limitations of existing CE methods to explain the decision-making process of complex ML models. UFCE allows for the inclusion of user constraints to determine the smallest set of feature modifications while considering feature dependence and evaluating the feasibility of suggested changes. Three experiments conducted using benchmark evaluation metrics demonstrated that UFCE outperformed two well-known CE methods regarding proximity, sparsity, and feasibility. The third experiment conducted on five datasets demonstrates the feasibility and robustness of UFCE on tabular datasets.  Furthermore, the results indicated that user constraints influence the generation of feasible CEs. Therefore, UFCE can be considered an effective and efficient approach for enriching ML models with accurate and practical CEs. The software and data are available as open source for the sake of open science at the Github repository of UFCE\footnote{\url{https://github.com/msnizami/UFCE}}.

In the present framework, UFCE adeptly manages binary classification problems. In forthcoming research endeavours, we intend to systematically extend our approach to encompass multi-class classification, thereby augmenting its suitability for a more extensive array of classification tasks.
The future work will extend the user involvement with a series of experiments (human-centered evaluations) to increase the usefulness of the developed framework. One of the prospects is human-grounded evaluations, which could be achieved by analysing the user's comprehension of the explanation. More specifically, we plan to design a cognitive framework for assessing the comprehension of explanations in a user study. 


\bibliographystyle{unsrtnat}
\bibliography{references}  






\end{document}